# Clinically Ready Magnetic Microrobots for Targeted Therapies

Fabian C. Landers[1]†, Lukas Hertle[1]†, Vitaly Pustovalov[1]†, Derick Sivakumaran[1,2]†, Oliver Brinkmann[1], Kirstin Meiners[3], Pascal Theiler[1], Valentin Gantenbein[1], Andrea Veciana[1], Michael Mattmann[1], Silas Riss[1], Simone Gervasoni[1,2], Christophe Chautems[1], Hao Ye[1], Semih Sevim[1], Andreas D. Flouris[4], Josep Puigmartí-Luis[5,6], Tiago Sotto Mayor[7,8], Pedro Alves[7,8], Tessa Lühmann[3], Xiangzhong Chen[9,10], Nicole Ochsenbein[11,12], Ueli Moehrlen[12,13], Philipp Gruber[14], Miriam Weisskopf[15], Quentin Boehler[1], Salvador Pané[1]*, Bradley J. Nelson[1]*

† Contributed equally; *Corresponding authors: vidalp@ethz.ch, bnelson@ethz.ch

[1] Multi-scale Robotics Laboratory, ETH. Zurich, Tannenstrasse 3, 8092 Zurich, Switzerland

[2] Magnebotix AG, Zurich, Switzerland

[3] Institute of Pharmacy and Food Chemistry, University of Würzburg, Am Hubland, 97074 Würzburg, (Germany)

[4] FAME Laboratory, University of Thessaly, Trikala, 42100, Greece

[5] Departament de Ciència dels Materials i Química Física, Institut de Química Teòrica i Computacional, University of Barcelona, Barcelona, Spain

[6] Institució Catalana de Recerca i Estudis Avançats (ICREA), Barcelona, Spain

[7] Transport Phenomena Research Centre (CEFT), Engineering Faculty, Porto University, Portugal

[8] Associate Laboratory in Chemical Engineering (ALICE), Engineering Faculty, Porto University, Portugal

[9] Institute of Optoelectronics, Shanghai Frontiers Science Research Base of Intelligent Optoelectronics and Perception, Fudan University, Shanghai 200438, People's Republic of China.

[10] Yiwu Research Institute of Fudan University, Yiwu 322000, Zhejiang, People's Republic of China

[11] Department of Obstetrics, University Hospital of Zurich, Rämistrasse 100, Zürich, 8092 Switzerland

[12] The Zurich Center for Fetal Diagnosis and Therapy, University of Zurich, Rämistrasse 71, Zürich, 8092 Switzerland

[13] Department of Pediatric Surgery, University Children's Hospital Zurich, Steinwiesstrasse 75, Zürich, 8092 Switzerland

[14] Kantonsspital Aarau AG, Institut für Radiologie/ Abteilung für diagnostische und interventionelle Radiologie, Tellstrasse 25, CH-5001 Aarau, Switzerland

[15] Center for Preclinical Development, University Hospital Zurich, University of Zurich, Zurich, Switzerland

**Abstract:**

Systemic drug administration often causes off-target effects limiting the efficacy of advanced therapies. Targeted drug delivery approaches increase local drug concentrations at the diseased site while minimizing systemic drug exposure. We present a magnetically guided microrobotic drug delivery system capable of precise navigation under physiological conditions. This platform integrates a clinical electromagnetic navigation system, a custom-designed release catheter, and a dissolvable capsule for accurate therapeutic delivery. *In vitro* tests showed precise navigation in human vasculature models, and *in vivo* experiments confirmed tracking under fluoroscopy and successful navigation in large animal models. The microrobot balances magnetic material concentration, contrast agent loading, and therapeutic drug capacity, enabling effective hosting of therapeutics despite the integration complexity of its components, offering a promising solution for precise targeted drug delivery.

**One-Sentence Summary:**

A complete microrobot system designed for transport and of targeted drug delivery to specific sites in the human body is presented.

**Introduction**

Severe side effects are often associated with systemic administration of drug treatments and are responsible for 30% of drug failures during clinical trials (*1*, *2*). In response to these challenges, magnetic micro- and nanorobots have been the focus of extensive research over the past two decades (*3–9*). These devices hold tremendous potential for targeted drug delivery, offering the possibility of transporting higher concentrations of therapeutic agents directly to disease sites, thereby enhancing treatment efficacy and minimizing side effects (*10*). During this time, significant advancements have been made in several key areas of microrobotics. Researchers have developed machines capable of sophisticated locomotion, with some designs mimicking or utilizing the movement of biological cells and microorganisms (*11*). These advances have been enabled by breakthroughs in novel materials (*12*), fabrication techniques (*13*), and processing approaches, which have greatly improved the control and functionality of these robots at micro and nanoscale levels (*14*).

Proof-of-concept studies have shown that these devices can be successfully loaded with various drugs, and early *in vivo* trials in small animals have validated their efficacy (*15–19*). In some cases, these microrobots have been enhanced with contrast agents, allowing for real-time tracking within the body (*20*, *21*). Despite this progress, the clinical translation of these technologies remains a significant challenge. A major obstacle is the fragmented nature of current research, where various aspects—such as locomotion, therapeutic loading, and imaging—have been developed separately without an integrated solution for clinical use.

For clinical use, microrobot materials must be highly biocompatible and ideally biodegradable to enable safe clearance from the body without invasive retrieval. Balancing biocompatibility, biodegradability, and the magnetic properties necessary for actuation presents a substantial constraint, restricting the overall achievable magnetization.

Another barrier to clinical adoption lies in the tools needed to precisely navigate these robots within the human body. Generating sufficient magnetic gradient strength for microrobot navigation across human-scale workspaces presents a considerable challenge, as magnetic field strength decays with the cube of distance from the source. While several magnetic navigation systems exist (*22–26*), most are too small or impractical for clinical settings. Our recent work introduced the Navion, an electromagnetic navigation system (eMNS) capable of generating magnetic fields across a clinically relevant workspace (*27*). For the navigation of mobile microrobots, precise control over both magnetic fields and gradients is essential due to their superparamagnetic nature.

For precise navigation and effective therapeutic delivery, real-time tracking is essential. The microrobot must be visible under clinically available imaging modalities to allow precise positioning during navigation. Additionally, the microrobot must achieve adequate drug loading to achieve therapeutic effects within target regions, coupled with a reliable release mechanism.

This paper presents a new type of microrobot that addresses these challenges by offering an integrated solution to meet the multiple requirements and constraints necessary for clinical application. Our platform combines a clinical eMNS, a release catheter, and a microrobotic capsule that acts as a microrobotic drug-carrying reservoir. The microrobot balances magnetic material concentration, contrast agent loading, and the capacity to carry therapeutic drugs in adequate concentrations, in which all capsule materials exhibit biocompatible properties and have received FDA approval for other intravenous applications. We demonstrated the clinical maturity of our robotic system through *in vitro*, *ex vivo*, and *in vivo* experiments. By addressing these longstanding issues, our work represents a significant step toward the practical deployment of magnetic microrobots in clinical settings, providing a flexible and scalable solution for advanced medical interventions.

## System Overview

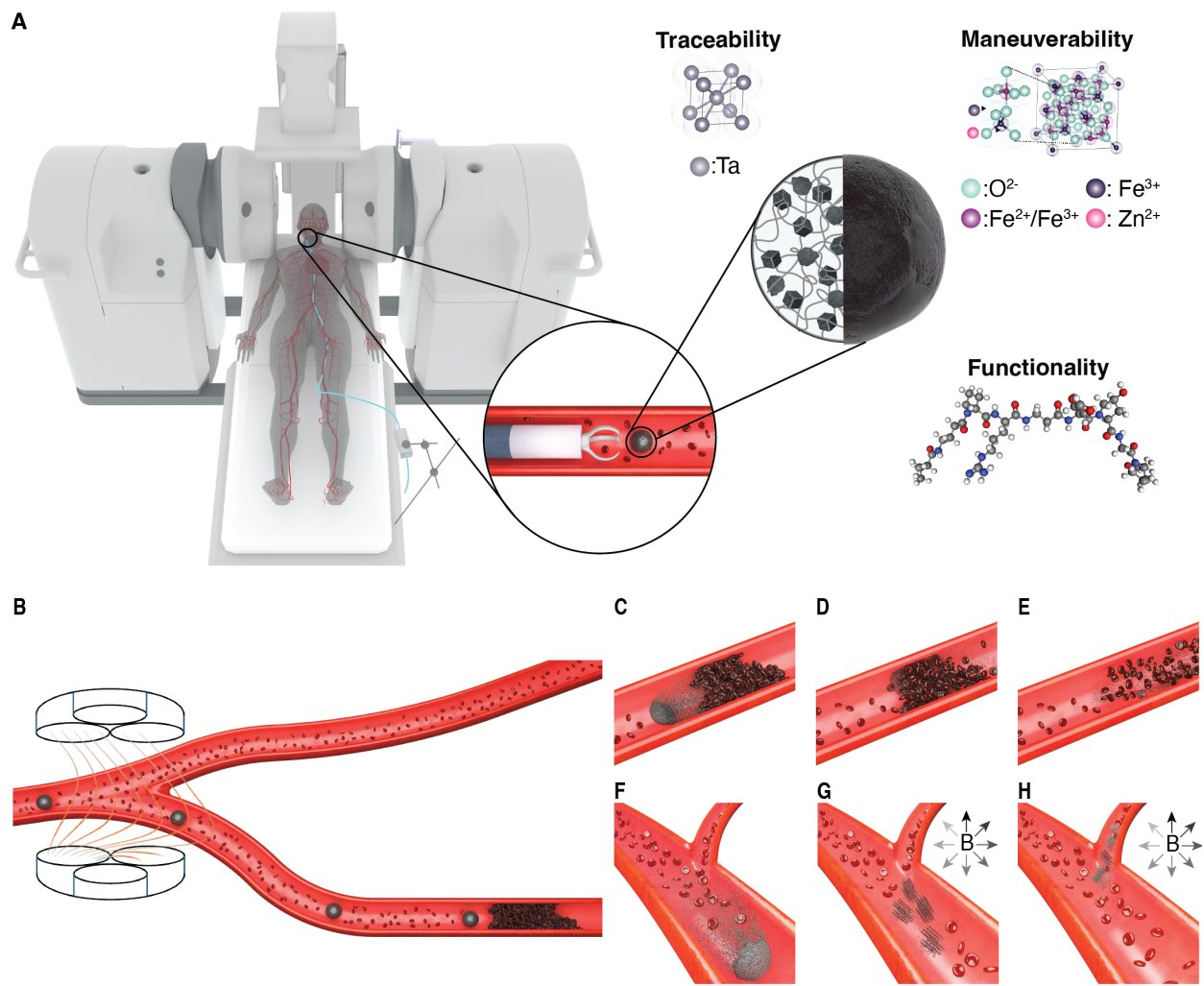

**Fig. 1. Magnetically guided microrobot platform for targeted drug delivery.** (A) Schematic of the robotic system within a clinical setup, featuring two electromagnetic navigation units, a custom-designed microrobot release catheter, and a magnetic microrobot. The microrobot is composed of a gelatin matrix embedded with zinc-doped iron oxide nanoparticles for magnetic actuation, tantalum nanoparticles for X-ray visibility, and therapeutic agents. (B) Schematic of microrobot navigation through the vascular network using magnetic gradient-based guidance. (C-E) Schematic of the thrombus dissolution process: (C) Microrobot dissolution is triggered via hyperthermia at the target site. (D) Initiation of clot lysis. (E) Restoration of vessel perfusion. (F-H) Schematic of nanoparticle chain formation and navigation into the microvasculature: (F) Hyperthermia-triggered microrobot dissolution at the target location. (G) Formation of nanoparticle chains using rotating magnetic fields. (H) Navigation of nanoparticle chain clusters into the distal microvasculature.

Reliable navigation under diverse physiological flow conditions is essential for achieving targeted drug delivery within human physiological fluids, such as blood or cerebrospinal fluid. In Fig. 1A, we conceptually illustrate our developed robotic system within a clinical setting. The system comprises a hierarchically structured modular arrangement, including two coupled electromagnetic units, a custom-designed release catheter, and a magnetically guidable untethered robotic end-effector, which we refer to as a magnetic microrobot or capsule. The modular design of our drug delivery platform enables seamless integration and adaptability to various clinical settings, leveraging existing infrastructure (e.g., fluoroscopes) and endoluminal surgery techniques.

Using a magnetic gradient-based in-flow navigation approach (Fig. 1B), magnetic gradients, or rotating magnetic fields, our untethered microrobot enables precise navigation within endovascular areas. The magnetic microrobot consists of a spherical gelatin matrix embedded with highly responsive magnetic zinc-doped iron oxide nanoparticles, radiopaque tantalum nanoparticles, and a therapeutic agent (Fig. 1A), all of which have been previously FDA-approved for various intravenous applications (*28–30*). This architecture is carefully designed to fulfill the requirements of magnetic navigation, X-ray trackability, biocompatibility, and biodegradability while allowing for the loading of therapeutic cargoes with different chemical natures. Through the application of high-frequency oscillating magnetic fields, iron oxide nanoparticles induce hyperthermic heating (*31*, *32*), triggering the dissolution of the microrobot and the precise release of its therapeutic payload (Fig. 1C-F). Once dissolved in low-flow regions, the nanoparticles can be actuated using rotating magnetic fields, forming dynamic magnetic nanoparticle chains that effectively transport the therapeutic agents into the distal microvasculature (Fig. 1G-H) (*33*).

**Robotic sub-systems**

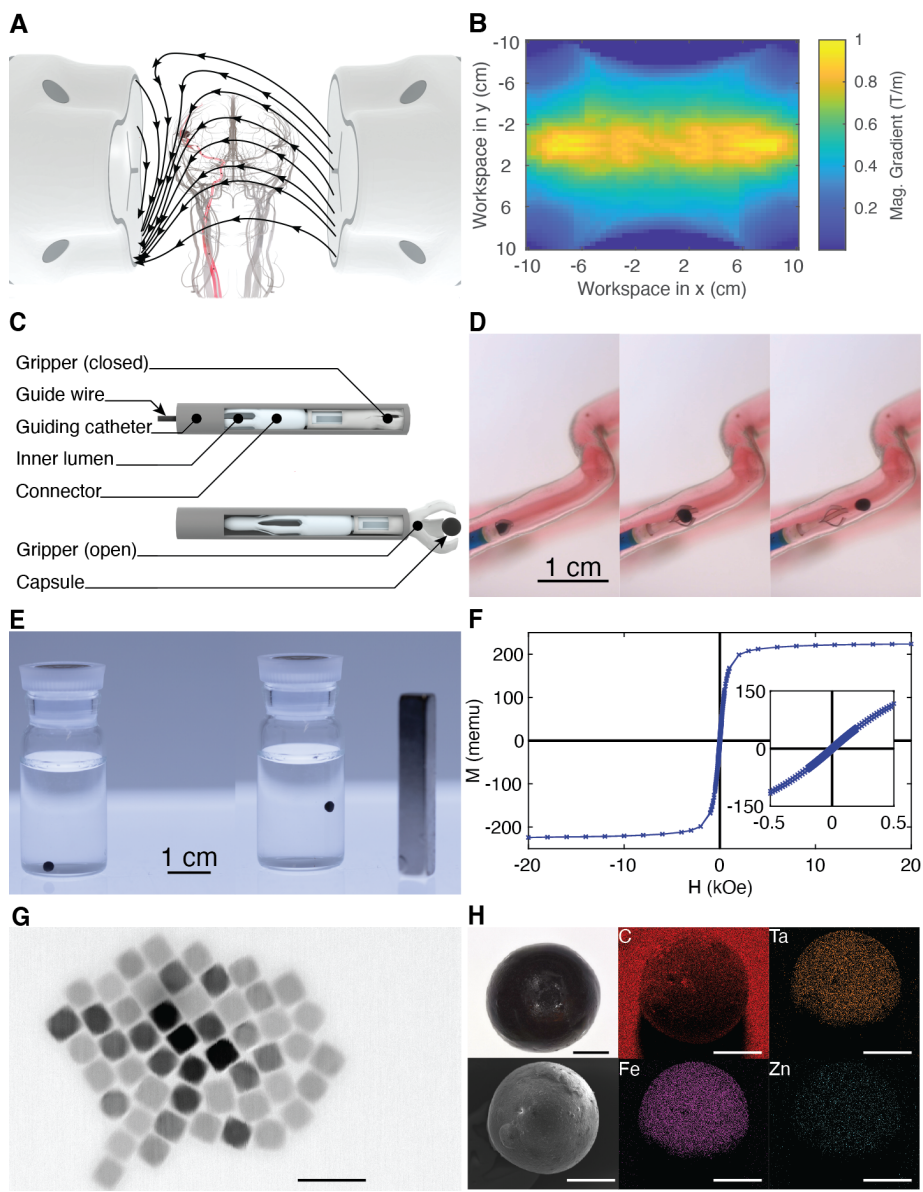

**Fig. 2. Robotic sub-systems for targeted drug delivery.** (A) Schematic of the electromagnetic navigation system (eMNS) illustrating the human cranial vasculature within the workspace generated by the dual Navion system. (B) Achievable magnetic field gradients at 30 mT in the x and y directions (with the z-direction fixed at the center) across a 20 cm × 20 cm workspace. (C) Custom release catheter design equipped with a flexible polymeric gripper for controlled microrobot deployment. (D) Sequential deployment of the microrobot from the catheter within a silicone vascular model under physiological conditions. (E) Microrobots in water, highlighting their spherical morphology and magnetic responsiveness. (F) Magnetic hysteresis loop of a microrobot, demonstrating magnetic properties at varying field strengths. (G) TEM image of synthesized Zn-substituted iron oxide nanocubes. Scale bar: 50 nm. (H) Optical microscope image (top left) and bright-field SEM image (bottom left) of a microrobot, with elemental EDX mapping verifying the presence of Zn-substituted iron oxide, tantalum for radiopacity, and carbon within the hydrogel matrix (middle and right). Scale bar: 500 μm.

We coupled two Navion systems (*27*), each featuring three electromagnetic coils separated by a distance of 35 cm to facilitate magnetic manipulation within a workspace large enough to accommodate a human skull. This double Navion system created heterogeneous magnetic fields and gradients within a 20 cm x 20 cm x 20 cm workspace (Fig. 2A and 2B), sufficient to accommodate a patient's head and a C-arm for fluorescent imaging. The system was designed to generate magnetic field gradients up to 1 T/m for magnetic fields of 30 mT (Fig. 2B).

To deliver the microrobot at the deployment site and avoid untethered navigation through regions of strong counterflow, our robotic system includes a first-stage release mechanism using a custom-built release catheter (Fig. 2C and 2D). The design uses a commercial catheter (7 Fr) with an inner guidewire connected to a flexible polymer gripper. When pushed beyond the outer guiding catheter, the polymer gripper opens and releases the pre-loaded microrobot (Fig. 2D, fig. S1A-D, and video S1). The gripper consists of high-density polyethylene (HDPE), meeting the specific load requirements for intravascular catheters outlined in ISO 10555 (*34*). Tensile stress testing demonstrated a maximum tensile load capacity of 16.07 N, aligning with the required peak tensile force specifications. Additionally, cyclic reliability testing of the flexible gripper tip demonstrated an elastic recovery of 77 % after ten deployments (fig. S1E).

To fabricate the magnetic capsule (Fig. 2E), we developed a microfluidic manufacturing approach using a custom-designed 3D flow-focusing chip (fig. S2A–C). This design facilitated the formation of a stable co-flow between the high-concentration hydrogel-nanoparticle suspension and the surrounding olive oil and Span® 80 surfactant mixture. Droplets of the hydrogel-nanoparticle suspension were dispensed into a cooled olive oil bath, enabling rapid hydrogel crosslinking during free-fall. This process resulted in the production of highly uniform spherical microrobots with tunable sizes. By adjusting flow rates, microcapsules with an average diameter of 1.69 mm (fig. S2D–E) were produced, optimized to fit the cranial vessels porcine models.

A sufficient magnetic force must be generated to enable effective maneuverability of the capsules within the human vasculature. This requires optimization of the scalar product of the magnetic gradient and the magnetization of the capsule (*35*). The capsule's magnetization is dependent on the magnetic susceptibility of the superparamagnetic iron oxide nanoparticles and the strength of the applied magnetic field. While higher magnetic fields enhance magnetization, they also limit the maximum achievable gradient. Considering the capabilities of the eMNS and balancing the magnetic field strength with the achievable magnetic gradients while considering the magnetic susceptibility of the capsule, a constant magnetic field of 30 mT was selected. The magnetization of the microrobot reached 72 memu at 30 mT (Fig. 2F), which is sufficient to lift the microrobot against gravity with the eMNS.

To achieve a high magnetic susceptibility of the capsules, we synthesized highly uniform zinc-substituted iron oxide nanocubes with an average edge length of 18.4 nm ± 1.3 nm, via thermal decomposition-based synthesis (Fig. 2G and fig. S3A-C) (*36*, *37*). These particles displayed an inverse spinel structure and excellent hyperthermic heating properties with a specific loss power (SLP) of 190 W/g (fig. S3D-E). The altered magnetic net spin of the particles resulted in an enhanced magnetic response at 30 mT and a high magnetic saturation of 96.9 emu/g (fig. S3F-G) (*38*). Before preparing the hydrogel-particle stock solution, the particles were functionalized with nitrodopamine (*39*), ensuring the formation of a stable colloidal dispersion (fig. S4). EDX mapping of the capsule confirmed the presence of Zn-substituted iron oxide, tantalum, and carbon within the capsule matrix (Fig. 2H).

**Maneuverability, traceability, and functionality of the microrobot**

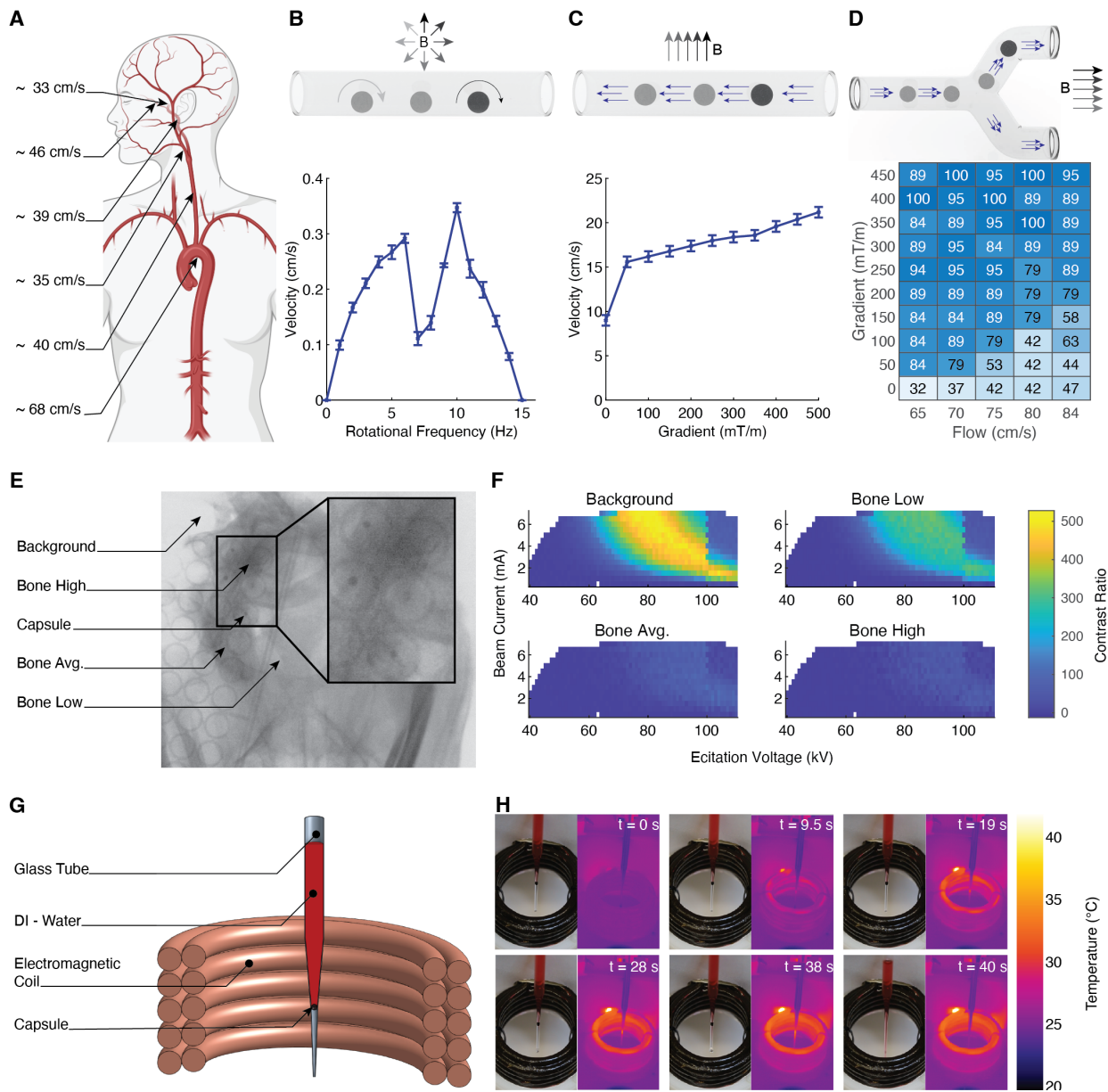

**Fig. 3. Capsule propulsion strategies, imaging, and dissolution mechanisms.** (A) Blood velocity map of the human arterial system with site-specific flow velocities indicated. (B) Microrobot propulsion via rolling along a liquid/solid interface under 10 mT rotating magnetic fields. (C) Gradient-based propulsion at 30 mT enabling microrobot movement against ambient flow. (D) In-flow navigation within a Y-shaped bifurcation, utilizing magnetic gradients at 30 mT to direct the microrobot into the desired branch. (E) Fluoroscopy imaging of microrobots beneath a porcine head *ex vivo*. (F) Contrast ratio maps of the microrobot in various regions of the porcine head, evaluated as a function of tube voltage (kV) and tube current (mA) of the X-ray tube. (G) Schematic representation of the hyperthermic dissolution process initiated by alternating magnetic fields at 510 kHz and 20 mT. (K) Thermal imaging showing the progressive heating and dissolution of the microrobot under high-frequency magnetic stimulation, with complete dissolution achieved within 40 seconds.

Blood velocities within the human arterial system vary drastically depending on the location (Fig. 3A) (*40*, *41*). To assess the maneuverability of the capsule at a 30 mT magnetic field under various vasculature flow conditions, we investigated three different propulsion approaches.

First, capsule rolling on a liquid/solid interface (e.g., vessel wall) induced by a rotating magnetic field achieved a peak forward velocity of 0.37 cm/s (Fig. 3B). While this velocity is insufficient for reliable navigation within the cardiovascular system, it is suitable for navigation within body regions with low flow velocities, such as the subarachnoid space. The forward velocity of the microcapsule increased with increasing rotational frequency of the magnetic field up to up to 5 Hz, followed by an intermediate decrease due to capsule bouncing. The forward velocity increased again for rotational frequencies between 7 and 10 Hz, after which the capsule's step-out frequency was reached (video S2).

Next, pulling the microrobot induced by magnetic field gradients enabled propulsion against flow velocities of up to 21.2 cm/s, with an almost linear relationship between the applied magnetic gradient and the flow velocity (Fig. 3C).

Under physiological conditions the encountered flows are higher than the propulsion exerted by gradient pulling on the microrobot. Thus, we also employed a navigation strategy where the flow naturally transports the microrobot. At the same time, the magnetic gradient steers the capsule into a desired trajectory, for example, a specific branch in the vasculature. We statistically analyzed this in-flow navigation within a Y-junction (Fig. 3D). The results showed a successful capsule delivery in more than 95% of the tested cases for flows up to 84 cm/s (video S3). We confirmed the viability of the in-flow approach through a numerical investigation using a computational fluid dynamics approach based on the finite-volume method (FVM; Supplementary text 1, figrs. S5-S9, and Table S1).

The traceability of the microrobot was confirmed through fluoroscopy imaging of the capsule *ex-vivo* under a porcine head using a mobile C-arm (Fig. 3E). A systematic contrast examination demonstrated the expected correlation between the X-ray tube power and the resulting contrast levels for various background contrast levels (Fig. 3F).

We tested the microrobot's ability to encapsulate different types of drugs despite incorporating multiple components. Specifically, we evaluated its drug-loading capacity and release profiles using doxorubicin (DOX) and ciprofloxacin (CIPRO). Additionally, we assessed the system's compatibility with biologics by incorporating tissue plasminogen activator (rtPA). rtPA displayed adsorption onto nanoparticles reduced enzymatic activity to 58%. Long-term rtPA stability studies revealed sufficient preservation of the single-chain form to comply with the European Pharmacopoeia (Supplemental text 2, fig. S10).

Continuous and slow drug release from the microrobot can be achieved through diffusion, which can be advantageous for conditions such as tumor treatment. In contrast, rapid, high-concentration release is preferred for acute therapies like thrombus dissolution. The microrobot's dissolution can be triggered by external magnetic stimulation, offering two key benefits: (1) control over the drug delivery rate at the target site and (2) a safety mechanism for immediate dissolution if the microrobot is navigated to an undesirable location. Under high-frequency magnetic stimulation (510 kHz at 20 mT), the magnetic nanoparticles incorporated in the microrobot exhibit hyperthermic heating, which initiates the dissolution of the thermo-responsive gelatin (Fig. 3G-H), and thus the release of the drug. The specific loss power of the particles was determined to be 190 W/g (fig. S3E), enabling rapid microrobot dissolution within 40 seconds.

**Cytotoxicity evaluation of the microrobot system**

The *in vitro* biocompatibility of the dissolved microrobots and its components were evaluated through cell viability tests. The cell viability of immortalized endothelial cells (EA.hy926), human mammary gland adenocarcinoma cells (SK-BR-3), and human embryonic kidney cells (HEK-293) was assessed at different particle concentrations (fig. S11A-C). To determine the influence of particle-hydrogel interactions on cell viability, we conducted assessments of the microrobot formulation (i.e., water 49 wt%, iron oxide 32 wt%, tantalum 28 wt%, gelatin 5 wt%). The viability of all cell types showed no statistically significant decrease for capsule concentrations up to 0.25 mg/mL compared to control groups. The viability of cells exposed to pure magnetic nanoparticles showed a notable decrease at lower concentrations, depending on the cell line, indicating a favorable interaction of gelatin-nanoparticle composites compared to pure nitrodopamine-functionalized nanoparticles.

The relevance of the developed approach for delivering drugs used for tumor treatment was assessed by loading DOX into the microrobots and analyzing in vitro cell viability after their dissolution. Endothelial cells (ECs) exhibited a similar decrease in viability when exposed to the same amount of pure DOX or DOX from loaded capsules, showing a significant decrease at concentrations of 0.2 µg/mL, compared to the control group exposed to non-drug-loaded capsules (fig. S11D-F). SK-BR-3 and HEK-293 cell lines were responsive to higher concentration of DOX concentrations. Additionally, microrobots were loaded with ciprofloxacin, with no significant increase in cell viability of SK-BR-3 and HEK-293 cells compared with pure ciprofloxacin. A decrease in cell viability was detected for ECs for capsule formulation concentrations higher than 0.5 mg/mL (fig. S11G-I).

# In vitro robotic system evaluation under realistic conditions

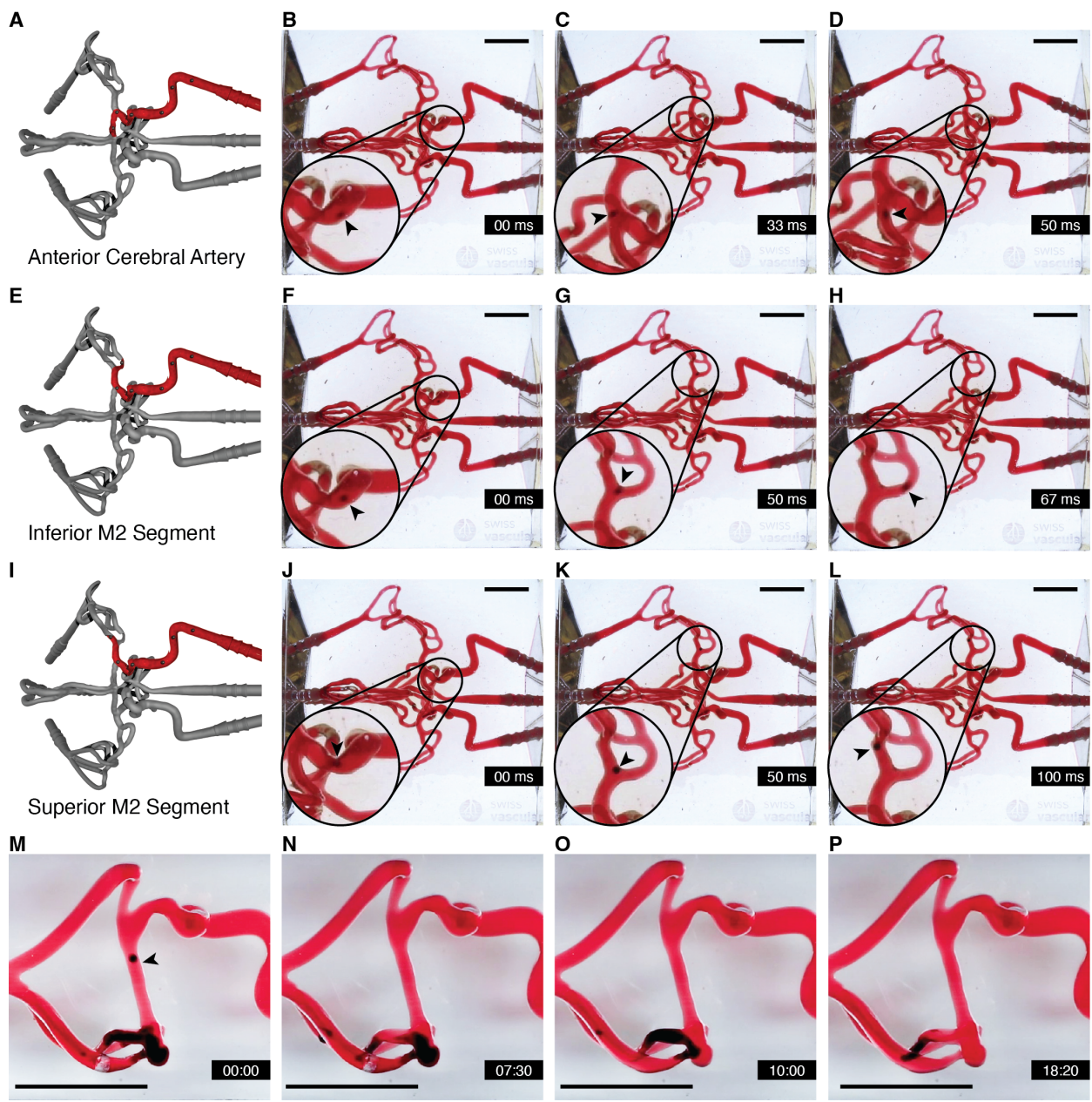

**Fig. 4. In-flow navigation and targeted drug delivery using microrobots in patient-specific vascular models.** (A-D) Navigation of a microrobot into the anterior cerebral artery (ACA) within 53 ms under a flow velocity of 37 cm/s in the internal carotid artery (ICA), utilizing a constant magnetic field of 30 mT and gradients up to 350 mT/m. (E-H) Navigation of a microrobot into the middle cerebral artery (MCA) and MCA M2 inferior trunk within 67 ms, under a flow velocity of 37 cm/s in the ICA, utilizing a constant magnetic field of 30 mT and gradients up to 350 mT/m. (I-L) Navigation of the microrobot into the MCA and MCA M2 superior trunk within 100 ms, under a flow velocity of 37 cm/s in the ICA, utilizing a constant magnetic field of 30 mT and gradients up to 350 mT/m. (M-P) Targeted delivery of an rtPA-loaded microrobot to a human blood clot occlusion in an MCA model under medium-flow conditions. Thermal dissolution

triggered rtPA and nanoparticle release, resulting in clot lysis within 7.5 minutes, with significant thrombus breakdown within 19 minutes. All Scale bars: 2.5 cm.

We evaluated the system's in-flow navigation capabilities in a three-dimensional silicone vascular model of a patient developed based on imaging data from magnetic resonance angiography (MRA). The microrobot was released into the internal carotid artery (ICA) under a flow velocity of 37 cm/s, typical in adult patients (*40*). The microrobot was successfully navigated into three distal vessels of the model's vasculature by utilizing a constant magnetic field of 30 mT and magnetic gradients up to 350 mT/m, in the desired directions.

First, a microrobot was steered into the anterior cerebral artery (ACA) within 53 ms (Fig. 4A-D). Next, a microrobot was directed into the Middle Cerebral Artery (MCA) and the MCA M2 inferior trunk within 67 ms (Fig. 4E-H). Lastly, a microrobot was again steered into the MCA and then into the MCA M2 superior trunk within 100 ms (Fig. 4I-L) (video S4). For enhanced visibility, we repeated the navigation of the microrobot in two M2 segments in a smaller model (MCA-aneurysm model) (video S5).

Targeted delivery of a complex therapeutic agent was demonstrated by guiding a rtPA-loaded microrobot to a human blood clot occlusion within the MCA model, under low-flow conditions. Thermal dissolution of the microrobot resulted in the release of rtPA and nanoparticles. Clot lysis was observed within 7.5 minutes from the release, resulting in most of the thrombus being broken down or flushed out within 19 minutes (Fig. 5M-P and video S6).

Utilizing a combination of thermal dissolution and low-frequency rotating magnetic fields, we further demonstrated the system's capability for microvasculature targeting (fig. S12A-D). An initial stage of localized embolization, utilizing multiple microrobots, effectively reduced the flow through the microvasculature. Nanoparticle chains were formed by applying low-frequency rotating magnetic fields, propelling the drug-loaded nanoparticles in- and against-flow-direction (fig. S12E-H).

## In vivo robotic system evaluation

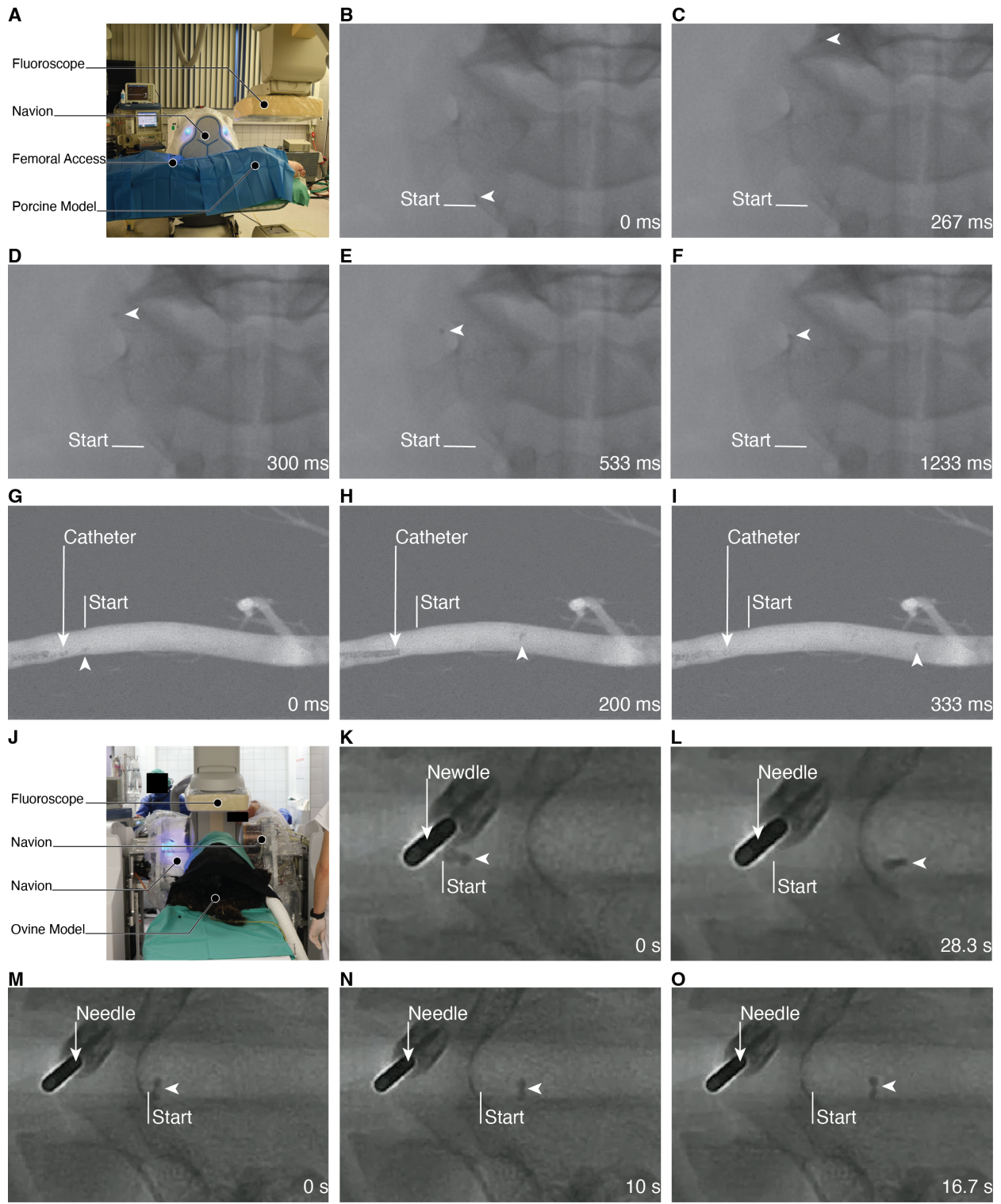

**Fig. 5. In vivo evaluation of the robotic platform in a clinical setting.** (A) Setup for *in vivo* experiments using a porcine model. (B-F) Fluoroscopic tracking of the microrobot released in the common carotid artery (CCA), navigating the cranial arteries within 1233 ms. (G-I) Real-time tracking of the microrobot within a pre-acquired vascular roadmap using digital subtraction

angiography (DSA) at 15 fps. (J) Experimental setup for evaluating microrobot navigation in the intrathecal space of an ovine model. (K-L) Navigation of microrobots into the fourth ventricle via the cisterna magna access using rotating magnetic fields. (M-O) Navigation of microrobots into the fourth ventricle via the cisterna magna access using magnetic gradients.

The clinical feasibility of the robotic platform was systematically evaluated through ex vivo and in vivo studies. Initially, we assessed the visibility and magnetic responsiveness of the capsules in an ex vivo model using the vasculature of a human placenta (Fig. S13).

Subsequently, we validated the fluoroscopic trackability of the microrobots *in vivo* in a porcine model under clinical conditions (Fig. 5A). Microrobots were released in the common carotid artery (CCA) and successfully tracked using a standard C-Arm fluoroscope at 30 frames per second (fps) as they navigated through the cranial arteries (Fig. 5BF, Video S7). Additionally, we demonstrated real-time tracking of the microrobot within a pre-acquired vascular roadmap using digital subtraction angiography (DSA) at 15 fps (Fig. 5G-I, Video S7).

To assess microrobot navigation in a complex anatomical region, experiments were conducted in an ovine model (Fig. 5J). Two microrobots (to enhance visualization of the navigation strategy) were released into the cisterna magna and magnetically guided into the fourth ventricle using a combination of rotating magnetic fields (Fig. 5K-L, Video S8) and gradient fields (Fig. 5M-N, Video S8). These experiments demonstrate the platform's capability to navigate microrobots through anatomically complex areas under clinical conditions, offering a minimally invasive alternative to current state-of-the-art surgical techniques.

**Discussion**

Our findings demonstrate the feasibility of navigating microrobots in vivo within large animal models under clinical conditions, marking an important step toward the clinical translation of microrobotic platforms. By integrating navigation, therapeutic delivery, and imaging into a unified system, this work addresses several of the longstanding challenges in the field of microrobotics. The use of materials that have been FDA-approved for other intravenous applications, coupled with the modular design of the robotic platform, ensures translatability and adaptability to a range of clinical workflows, providing a foundation for future applications in minimally invasive medicine.

Targeted drug delivery offers a promising alternative to systemic administration by enabling localized therapeutic administration, reducing off-target effects, and potentially improving treatment efficacy. Our system demonstrated precise navigation of microrobots under realistic physiological conditions, effective drug loading and release mechanisms, and compatibility with standard imaging modalities. These features suggest that microrobots could be employed in a variety of therapeutic contexts, including the treatment of vascular occlusions, localized infections, or tumors.

The successful *in vivo* visualization and manipulation in porcine and ovine models represents a significant milestone for microrobotic drug delivery, as this is the first time that microrobots have been demonstrated to operate effectively within a clinically relevant workspace and flow environment. Additionally, targeted thrombolytic therapy in an *in vitro* vascular occlusion model highlights the potential for using microrobots in acute interventions, such as ischemic stroke treatment. Our experiments in an ovine model further validate the platform's ability to navigate through anatomically complex regions, such as the central nervous system, where conventional surgical approaches are highly invasive.

Future efforts should prioritize long-term safety studies, scalability, and the development of automated navigation algorithms to minimize operator dependency and enhance the system's clinical applicability.

In summary, this study establishes a new benchmark for microrobotic systems, showcasing their potential for precise, minimally invasive therapeutic interventions. While significant work remains to fully translate this technology into clinical practice, our results provide a robust framework for addressing the complex challenges associated with targeted drug delivery.

**Acknowledgments:**

This project has received funding from the European Union's Horizon 2020 Proactive Open program under grant agreement No 952152. This work was also supported by the Swiss National Science Foundation through grant number 200020_212885, and the ITC-InnoHK grant 16312.

The authors acknowledge the support of the FIRST Center for Micro- and Nanoscience and the Scientific Center for Electron Microscopy (ScopeM) of the Swiss Federal Institute of Technology (ETHZ).

**Conflict of interests**

C.C. and B.J.N. are co-founders of NanoFlex Robotics AG. B.J.N. is a co-founder of Magnebotix AG. F.C.L., O.B., P.T., S.P., and B.J.N. are co-founders of Swiss Vascular GmbH

# Supplementary Material

## Clinically Ready Magnetic Microrobots for Targeted Therapies

Fabian C. Landers[1]†, Lukas Hertle[1]†, Vitaly Pustovalov[1]†, Derick Sivakumaran[1,2]†, Oliver Brinkmann[1], Kirstin Meiners[3], Pascal Theiler[1], Valentin Gantenbein[1], Andrea Veciana[1], Michael Mattmann[1], Silas Riss[1], Simone Gervasoni[1,2], Christophe Chautems[1], Hao Ye[1], Semih Sevim[1], Andreas D. Flouris[4], Josep Puigmartí-Luis[5,6], Tiago Sotto Mayor[7,8], Pedro Alves[7,8], Tessa Lühmann[3], Xiangzhong Chen[9,10], Nicole Ochsenbein[11,12], Ueli Moehrlen[12,13], Philipp Gruber[14], Miriam Weisskopf[15], Quentin Boehler[1], Salvador Pané[1]*, Bradley J. Nelson[1]*

† Contributed equally; *Corresponding authors: vidalp@ethz.ch, bnelson@ethz.ch

[1] Multi-scale Robotics Laboratory, ETH. Zurich, Tannenstrasse 3, 8092 Zurich, Switzerland

[2] Magnebotix AG, Zurich, Switzerland

[3] Institute of Pharmacy and Food Chemistry, University of Würzburg, Am Hubland, 97074 Würzburg, (Germany)

[4] FAME Laboratory, University of Thessaly, Trikala, 42100, Greece

[5] Departament de Ciència dels Materials i Química Física, Institut de Química Teòrica i Computacional, University of Barcelona, Barcelona, Spain

[6] Institució Catalana de Recerca i Estudis Avançats (ICREA), Barcelona, Spain

[7] Transport Phenomena Research Centre (CEFT), Engineering Faculty, Porto University, Portugal

[8] Associate Laboratory in Chemical Engineering (ALICE), Engineering Faculty, Porto University, Portugal

[9] Institute of Optoelectronics, Shanghai Frontiers Science Research Base of Intelligent Optoelectronics and Perception, Fudan University, Shanghai 200438, People's Republic of China.

[10] Yiwu Research Institute of Fudan University, Yiwu 322000, Zhejiang, People's Republic of China

[11] Department of Obstetrics, University Hospital of Zurich, Rämistrasse 100, Zürich, 8092 Switzerland

[12] The Zurich Center for Fetal Diagnosis and Therapy, University of Zurich, Rämistrasse 71, Zürich, 8092 Switzerland

[13] Department of Pediatric Surgery, University Children's Hospital Zurich, Steinwiesstrasse 75, Zürich, 8092 Switzerland

[14] Kantonsspital Aarau AG, Institut für Radiologie/ Abteilung für diagnostische und interventionelle Radiologie, Tellstrasse 25, CH-5001 Aarau, Switzerland

[15] Center for Preclinical Development, University Hospital Zurich, University of Zurich, Zurich, Switzerland

## Materials and Methods

***Synthesis of Nitrodopamine*** Nitrodopamine hydrogensulfate was synthesized following a previously reported protocol.(*42*) In short, 21 mmol of dopamine hydrochloride (Sigma Aldrich) was dissolved in 120 mL de-ionized water before adding 72.4 mmol sodium nitrite (Sigma Alrich) and cooling the mixture to 0°C. Following, 40 ml of 20 vol/vol% of sulfuric acid (Sigma Aldrich) were added dropwise under constant stirring. After addition, the reaction was allowed to proceed at room temperature for 16 hours. The resulting yellow crude product was collected by vacuum filtration and washed multiple times with de-ionized water (4 °C). Finally, the resulting nitrodopamine hydrogensulfate was collected, dried under a high vacuum, and stored at 4 °C for further use.

***Synthesis of zinc ferrite nanoparticles*** Cubic zinc-substituted iron oxide nanoparticles were synthesized *via* a previously established thermal-decomposition protocol with multiple modifications.(*43*) In a typical synthesis, first 1.52 mmol of iron(III) acetylacetonate (Thermo Fisher Scientific), 0.23 mmol of zinc(II) acetylacetonate (Sigma Aldrich), and 1.31 mmol of sodium oleate (Tokyo Chemical Company) were dissolved in a mixture of 7 ml benzyl ether (Thermo Fisher Scientific), 15 ml of 1-octadecene (Sigma Aldrich), 3 ml of tetradecene (Sigma Aldrich), and 5.25 mmol of oleic acid (Sigma Aldrich). The mixture was then transferred into a three-bottle neck flask equipped with a thermocouple and reflux condenser before being degassed at 60 °C for 60 minutes. Afterwards the solution was heated to reflux at 294 °C (3°C min$^{-1}$) under a constant $N_2$ flow at which it was kept for 90 minutes. The resulting particle containing crude product was eventually cool to room temperature before collection and stored under $N_2$ at -20 °C for further processing.

***Ligand modifications of nanoparticles*** Ligand exchange of oleic acid-covered nanoparticles with nitrodopamine: To achieve high colloidal stability of the particles in aqueous media, a single-step ligand exchange with nitrodopamine was undertaken following a previously established procedure with small adjustments.(*42*) As-synthesized particles were first purified by multiple sequential redispersion and sedimentation processes. First, 240 ml of particle-containing stock solution was mixed with 240 ml of chloroform (Sigma Aldrich) and 720 ml of acetone (Sigma Aldrich), followed by sedimentation *via* centrifugation. Next, the particles were washed three times by redispersing them in 250 ml chloroform (Sigma Aldrich) and sedimentation after adding 750 ml of a 50:50 vol:vol methanol (Sigma Aldrich) and acetone (Sigma Aldrich) solution. Finally, the particles were washed again with a mixture of 250 ml chloroform and 750 ml acetone (Sigma Aldrich) before redispersion in 40 ml THF anhydrous (Sigma Aldrich). Ligand exchange with nitrodopamine was subsequently undertaken by dissolving 400 mg of nitrodopamine in 10 dry dimethylformamide (DMF) (Sigma Aldrich) before adding 30 ml of THF anhydrous (Sigma Aldrich) within a nitrogen-filled glovebox. Next, the solution was heated to 40 °C before dropwise addition of 40 ml particle-containing THF under vigorous stirring. The ligand exchange was allowed to proceed for 4 hours under inert conditions before sedimentation of particles functionalized with nitrodopamine in 500 ml of cold acetone (Sigma Aldrich). Functionalized particles were washed four times by redispersion and centrifugation in 400 ml methanol (Sigma Aldrich) and four times by redispersion and sedimentation in 400 ml de-oxygenated de-ionized water before collection. *Nitrodopamine functionalization of tantalum nanoparticles*: Commercial Ta nanoparticles (tantalum powder, 500 nm – 1 µm, GoodFellow) were functionalized with nitrodopamine by dissolving 50 mg nitrodopamine in 50 ml de-oxygenated de-ionized water containing 500 mg of Ta particles. The solution was subsequently sonicated for 1 hour, allowed to stand at room temperature for 16 hours, and sonicated once more for 1 hour before particle sedimentation in 200 ml cold acetone (Sigma Aldrich). Nitrodopamine-functionalized particles were washed four times in 200 ml of methanol (Sigma Aldrich) and four times in 200 ml de-oxygenated de-ionized water before collection.

***Microscopy***
*Light Microscopy:* Microscope images of millimeter- and micrometer-sized structures were obtained with a KEYENCE VHX-X1 optical microscope (Keyence International).
*Scanning Electron Microscopy (SEM):* SEM images were recorded with a Zeiss ULTRA 66 (Carls Zeiss GmbH) operating at 5 kV. EDX mapping analysis was undertaken with a 60 µm aperture and 10 mm working distance at 20 kV.
*Transmission electron Microscopy (TEM):* TEM, scanning TEM, and EDX of nanoparticles were undertaken with an FEI Talos F200X (Chem S/TEM) (Thermo Fisher Scientific) operating at 200 kV. Samples were prepared by drop-casting diluted particle dispersions on a carbon-coated Cu TEM grid (400 mesh) (1824, Ted Pella).

***Structural analysis***

*Fourier Transform Microscopy (FTIR):* Spectra of samples were recorded with a Bruker Tensor 27 spectrometer (Bruker Corp.) across a range from 4000-400 cm$^{-1}$ with a resolution of 4 cm$^{-1}$.

*Nuclear Magnetic Resonance (NMR) Spectroscopy:* $^1$H-NMR spectra were collected by a 400 MHz Bruker Advance Ultrashield (Bruker Corp.), using DMSO-d6 as solvent.

*Thermogravimetric Analysis (TGA):* Thermogravimetric measurements on dry samples were taken with a Metter Toledo TGA/DSC 3+ Star up (Mettler-Toledo) under a steady O$_2$ flow of 60 ml min$^{-1}$ from 30°C to 900°C (10°C min$^{-1}$).

*X-ray Diffraction (XRD) analysis:* The crystal structure of nanoparticle powders was derived with a Malvern Panalytical Empyrean diffractometer (Malvern Panalytical GmbH) equipped with a copper X-ray source ($\lambda$ = 1.5406 Å) and a PIXcel detector. Measurements were carried out within the range of 4° ≤ 2θ ≤ 80° with a step size of 0.04° and a sweeping rate of 0.2 sec per step. Using the Scherrer equation, crystallite sizes were calculated from the main Bragg peak with the highest intensity.

*Direct Light Scattering (DLS) & ζ Potential measurements:* Hydrodynamic diameters and zeta potentials of diluted particle dispersions in de-ionized water were measured with a Malvern Panalytical Zetasizer Nano ZS (Malvern Panalytical GmbH) at 25°C.

*Specific Loss Power (SLP) measurements:* Heating capacities of diluted dispersions of phase-transferred nanoparticles were evaluated under AC magnetic field stimulation at 510 kHz and 20 mT (NAN201003 Magnetherm, nanoTherics Ltd.). SLP values were calculated from recorded temperature curves using the formula $\boldsymbol{SLP = \frac{C}{m}\frac{dT}{dt}}$, in which $\boldsymbol{C}$ refers to the heat capacity of water per unit volume (4.184 J K$^{-1}$ mL$^{-1}$), $\boldsymbol{m}$ to the concentration of nanoparticles per volume water (g mL$^{-1}$) and $\frac{dT}{dt}$ is the experientially recorded temperature change per unit time (°C sec$^{-1}$).

**Magnetic characterization** Symmetric magnetic hysteresis loops were measured for nanoparticles and capsules over the field ranges of 0.1, 0.2, 0.3, 0.4, and 2 T at 300 K, using a vibrating sample magnetometer (VSM EZ9, Microsense). Nanoparticle samples were prepared by drop casting highly concentrated dispersions of oleic acid respectively nitrodopamine-functionalized particles onto a round filter-paper substrate (Ø: 8mm) and drying under a high vacuum for 48 hours. Normalization to magnetization per gram of zinc ferrite was undertaken by accounting for organic weight percentage loss determined *via* TGA.

**3D microfluidic chip fabrication** Polytetrafluoroethylene (PTFE) tubes with an inner diameter of 0.8 mm and an outer diameter of 1.6 mm (Schlauch PTFE 0.8x1.6 mm, Semadeni) were placed on a 18 G blunt-end Luer-lock dispensing needles (Stainless Steel Dispensing Needles with Luer Lock Connection, McMaster-Carr) and inside a 3D printed (Phrozen Sonic mini 8K resin printer, Prozen Aqua 8K resin) rectangular mold. Next, polydimethylsiloxane (PDMS, Sylgard 184, Dow Croning) was filled into the mold, degassed under vacuum, and cured for 2 hours at 80 °C. After cooling to room temperature, 3D microfluidic chips were cut out of the mold, and the PTFE tube was removed. An additional access to the chip was punched perpendicularly to the dispensing needle using a 1.5 mm microfluidic punch (Rapid-Core Microfluidic Punch, 1.5 mm, Darwin Microfluidics). Lastly, the dispensing needle was removed from the chip and replaced with an 18 G blunt-end lure lock needle (Blunt-end Luer Lock Syringe Needles, 18 G, Darwin Microfluidics).

**Fabrication of magnetic capsule microrobots** A magnetic hydrogel formulation comprising 5 wt% gelatin, 37 wt% nitrodopamine-functionalized magnetic nanoparticles, and 16 wt% nitrodopamine-functionalized Ta nanoparticles was prepared by dispersing the respective amounts of the individual components within 1ml of de-ionized water at 55 °C under a combination of mechanical stirring and ultrasound. The hydrogel formulation and olive oil (Sigma Aldrich) with 2 % Span-80 (Sigma Aldrich) were loaded into 5 ml and 10 ml glass syringes, respectively. Next, the syringes were loaded into a syringe pump (Nemesis, Cetoni GmbH) and heated to 55 °C *via* syringe heating pads (New Era). The outlets of the syringes were connected to the 3D microfluidic chip, and a hydrogel/oil co-flow was achieved by pumping the olive oil at a flow rate of 200 µl/min and the hydrogel at 100 µl/min. A PVC tubing (inner diameter 1 mm, outer diameter 2 mm, McMaster Carr) was connected to the 3D microfluidic chip. The droplets forming at the outlet were collected in a cooled (4 °C) olive oil bath. After fabrication, the capsules were cooled down in the oil bath for 15 minutes and were transferred to a fridge (5 °C) for at least 12 hours. Optionally, the capsules were dried in air for two hours and stored at 5°C.

**Fabrication of rtPA loaded capsules** rtPA loaded capsules were fabricated following the protocol described above (*Fabrication of magnetic capsule microrobots*) and by adding 9 mg/ml recombinant tissue-type plasminogen activator (rtPA) in form of 427 mg/ml powder formulation (Actilyse® 50 mg, Boehringer Ingelheim) to the hydrogel formulation before loading it into the glass syringe.

***Small molecule drug loading:*** Doxorubicin and ciprofloxacin, two small-molecule drugs, were loaded into the capsules through an incubation process. Freshly prepared capsules were stored overnight in olive oil, followed by incubation in stock solutions of 18.4 mM doxorubicin or 30.2 mM ciprofloxacin for 96 hours. After incubation, the drug-loaded capsules were dried and washed three times with deionized water.

***Small molecule drug release kinetic assessment*** Drug release kinetics of doxorubicin and ciprofloxacin were assessed by submerging drug-loaded capsules in phosphate-buffered saline solutions at pH 7.4 and 6. Samples were kept at 37°C under mild agitation, and sample aliquots were taken after predetermined time steps. Sample aliquots were centrifuged at 14000 rpm for 30 min. Then, the supernatant was collected and centrifuged again at 14000 rpm for 15 min. Finally, 100 µL of supernatant was collected for absorbance scans. Doxorubicin and Ciprofloxacin releases were recorded with absorbance measurements at the wavelength of 480 nm and 316 nm (Infinite 200 Pro, Tecan).

***Assessment of rtPA stability, activity, and compatibility*** Recombinant tissue-type plasminogen activator (rtPA) was purchased as lyophilized powder formulation (Actilyse® 50 mg, alteplase for injection E.P., PZN 03300636) from Boehringer Ingelheim (Ingelheim am Rhein, Germany). According to the drug manufacturer, 1 mg rtPA corresponds to 46.66 mg Actilyse® powder formulation and 580,000 International Units (IU). Fluorescence substrate glutaryl-L-glycine-L-arginine-7-amino-4-methylcumarin-hydrochloride (Glutaryl-Gly-Arg-AMC, > 99%) was purchased by Bachem AG (Bubendorf, Switzerland). Chromogenic substrate methanesulfonyl-D-hexahydrotyrosine-L-glycine-L-arginine-p-nitroanilide-acetate (T2943, ≥ 95%) was purchased from Merck KGaA (Darmstadt, Germany).

*Determination of proteolytic activity of tPA by fluorescence activity* The enzyme activity of rtPA was measured as previously described.(*44*) Briefly, 100 µM prewarmed glutaryl-Gly-Arg-AMC substrate dissolved in 0.1 M Tris-HCl buffer pH 8.5 containing 0.02% Tween 80 was mixed with the rtPA samples. After incubation at 37°C for 20 minutes, the reaction was stopped by adding 2% SDS solution. Fluorescence intensity of proteolytically cleaved AMC was measured at emission and excitation wavelengths of 455 and 383 nm at 37°C in Tecan Reader Infinite M Plex (Tecan Group, Maennedorf, Switzerland). Measurements were performed in triplicate.

*Determination of single-chain rtPA content by size exclusion chromatography* Single-chain rtPA (sc-rtPA) content was analyzed via size exclusion chromatography as described in the European Pharmacopoeia monograph for Alteplase for injection.(*45*) Agilent 1260 infinity II HPLC (Agilent Technologies Inc., Waldbronn, Germany) with a TSKgel G300SW 10 µm 600 ×7,5 mm size exclusion (SEC) column (Tosoh Bioscience GmbH, Griesheim, Germany) was equipped with a variable wavelength detector (G7115A, Agilent), an automatic vial sampler (G7129C, Agilent), a pump (G7104C, Agilent), and a multicolumn oven (G7116A, Agilent). The mobile phase was 30 g/L sodium dihydrogen phosphate and 1 g/L sodium dodecyl sulfate (SDS) pH 6.8 in Millipore water with an isocratic flow of 0.5 mL/min for 60 minutes. The injection volume was 50 µL and the detector was set to a wavelength of 214 nm. The samples were prepared by incubating one part test solution with three parts of 3 g/L dithiothreitol in the mobile phase at 80°C for 4 minutes.

*Stability of rtPA over three months*
The stability of rtPA was investigated by storing 1 mg/mL rtPA solutions in water for injection or 5% (m/V) gelatin solution at 5°C or 25°C for three months. The change in proteolytic activity was determined by fluorescence activity assay. The change in sc-rtPA content was determined for the water for injection samples by size exclusion chromatography. Samples were stored in triplicate for each time point, and activity measurements were performed in triplicate.

*Interaction of rtPA with iron oxide nanoparticle surface* The interaction of rtPA with iron oxide nanoparticles as investigated by sodium dodecyl sulfate-polyacrylamide gel (SDS-Page) electrophoresis under reductive conditions. For this, 1 mg/mL rtPA solution was mixed with nanoparticle solution or nanoparticle supernatant and incubated for one hour. The samples were centrifuged and mixed with solubilization buffer (250 mM Tris-HCl pH 8.0, 7.5% w/v SDS, 25% v/v glycerol, 0.25 mg/mL bromophenol blue, 12.5% v/v β-mercaptoethanol) and incubated at 95°C for 5 minutes. SDS-Page was performed using 37.5 mM Tris-HCl with 1 g/L SDS for stacking (pH 6.8) and separating gel (pH 8.8), the latter with 12% acrylamide. 25 mM Tris-HCl with 192 mM glycine and 1 g/L SDS was used as a running buffer. Staining was performed with Coomassie blue G-250 solution.
The stability of rtPA with nitro dopamine-coated iron oxide nanoparticles was investigated by incubating 0.1 mg/mL (60000 IU/mL) rtPA solution with 7.5 mg/mL iron oxide nanoparticle solution for one hour at room temperature. The residual enzyme activity of the NP-rtPA mixture was measured by fluorescence activity assay as described above. In addition, the residual enzyme activity of the supernatant of this mixture after centrifugation was measured. Measurements were performed in triplicate.

***Cytotoxicity/Cell viability***

Dulbecco's Modified Eagle's high glucose Medium and nuclease-free buffered saline pH 7.4 (PBS) were purchased from Merck KGaA (Darmstadt, Germany). Fetal bovine serum (FBS) and MEM non-essential amino acids solution (100x) were purchased from ThermoFisher Scientific (Karlsruhe, Germany). WST-1 was purchased from Roche Diagnostics GmbH (Mannheim, Germany).

*Cell culture:* Human embryonic kidney cells (HEK293; ATCC, CRL-1573), human mammary gland adeno carcinoma cells (SK-BR-3; ATCC, HTB-30), and human immortalized endothelial cells (EA.hy926; ATCC, CRL-2922) were cultured in Dulbecco's modified Eagle's medium (DMEM) supplemented with 10% (v/v) FBS, 100 U/mL penicillin G, and 100 µg/mL streptomycin. EA.hy926 cell line was further supplemented by 1% (v/v) non-essential amino acids (Thermo). All cells were cultured in humidified 5% $CO_2$ at 37°C. Passages 8-10 were used for the experiments.

*Cytotoxicity:* The cytotoxicity of the capsule ingredients, empty capsules, and capsules loaded with Doxorubicin or Ciprofloxacin were examined using HEK293 cells, SK-BR-3 cells, and EA.hy926 cells, seeded in a 96-well plate format (10000 cells/well; 100 µL per well) in the respective cell medium. The cells were treated with a dilution series of the capsules of 0.50 to 0.0020 mg capsule/mL in phosphate-buffered saline pH 7.4. The investigated capsule components were examined in the concentration range corresponding to the concentration of the capsule components in the dilution series of the capsules. (iron oxide nanoparticles: 0.33 to 0.0013 mg/mL; tantalum powder: 0.13 to 0.00053 mg/mL) After 24 hours of treatment, cells were incubated with WST-1 for 3 hours at 37°C according to the manufacturer's instructions. Prior to measurement, the 96-well plates were centrifuged for one minute (4000 rpm), and the supernatants were transferred to new plates. The mitochondrial activity of the cells after incubation with the capsules and capsule ingredients was determined by measuring the absorbance of the soluble formazan product in the supernatant at 450 nm as well as background noise at 630 nm using Tecan Reader Infinite M Plex. Statistical analysis was performed with OriginPro 2020 (OriginLab Corporation, Northampton, USA). A comparison of cell viability was performed with analysis of variance (ANOVA). Samples with p-value <0.05 were considered statistically significant and were marked with an asterisk (*).

### *In vitro navigation*

All navigation tests were performed in water at room temperature. For rolling and in-flow navigations, 0.04 vol% of red food coloring (Bordeaux food color, Trawosa AG) was added for better visibility.

*Rolling:* Rolling navigation was performed by placing capsules inside a tube (6.3 mm inner diameter, polyurathene rubber, McMaster Carr). A rotating magnetic field of 10 mT and various rotational frequencies between 0 Hz and 15 Hz was generated using a Navion system.(*46*) All experiments were video recorded and the velocities were calculated.

*Against flow:* Characterization of the motion against flow were performed by placing capsules inside a tube (6.3mm inner diameter, polyurathene rubber). Magnetic gradients between 0 mT/m and 500 mT/m countering the flow direction and a constant magnetic field of 30 mT were applied using an OctoMag electromagnetic navigation system (Magnebotix AG). A continuous water flow was applied using a custom-built peristaltic system (consisting of two peristaltic pumps with a 90° phase shift to reduce pulsatility), opposing to the direction of the magnetic gradient. The flow was increased until the capsule was flushed out of the workspace and the final flow was recorded.

*In-flow:* In-flow maneuverability as a function of the applied magnetic field gradient strength and flow speed was assessed in a custom molded silicon Y-junction (90°). In a typical characterization 19 capsules were flushed through the Y-junction at predefined flow speed velocities while applying a magnetic gradient perpendicular to the flow direction towards the targeted Y-junction branch and a constant magnetic field of 30 mT. The electromagnetic navigation system consisted of two Navions (NanoFlex Robotics AG and Magnebotix AG) facing each other. Maneuverability was subsequently assessed by calculating the percentage of successfully actuated capsules.

*3D navigation: In vitro* three-dimensional navigation was demonstrated within a patient data-based silicone vascular model (Swiss Vascular GmbH). The fluid flow entering the internal artery (ICA) and basilar artery (BA) was adjusted to 37 cm/s and 35 cm/s, using a flow sensor (SLF3S-400B, Sensirion).

*Targeting of micro-vasculature:* Microvasculature embolization and locomotion experiments were conducted in a *Rete Mirabile* model (Swiss Vascular GmbH). All experiments were performed under a flow rate of 10 cm/s, using water with 0.04 vol% red food coloring (Bordeaux food color, Trawosa AG) and the OctoMag (Magnebotix AG) electromagnetic navigation system. Initially, multiple capsules were magnetically guided in-flow towards the *Rete Mirabile* structure. Capsule dissolution was triggered by raising the water temperature to 40 °C for 30 seconds. Successful embolization and significant flow reduction were demonstrated by navigating nanoparticle swarms both in-flow and against the flow direction using a rotating magnetic field at 1 Hz and 30 mT.

**Fabrication of capsule release catheter:** *Laser milling:* High-density polyethylene (HDPE) medical tubes (inner diameter: 0.0287 in, outer diameter: 0.0335 in) were cut using a scalpel.

*Mold 3D printing:* Molds to thermoform the release catheter's capsule release grippers were designed using computer aided design software (SolidWorks) and processed with slicer software (LycheeSliver v5.4.3) into G-code format. Next, the molds were 3D printed using a stereolithography 3D printer (Phrozen Sonic mini 8K) with a high-resolution photo resin (Aqua 4K, Phrozen). The 3D prints were cleaned in isopropanol (Sigma Aldrich) for 15 minutes and UV post-cured at 40 °C for 15 minutes (Form Cure , Formlabs).
*Thermoforming:* Shaping of the individual digits was performed by placing the pre-cut HDPE tubes inside the 3D-printed mold and placing it in an oven at 130 °C for 20 minutes.
*Catheter assembly:* The release catheter tip was glued onto a 3D printed mount and commercial guide wire using a UV-curable adhesive (Adhesive 68, Norland optical).

***Ex vivo human placenta study:*** A human placenta was donated with informed written consent and approval from the Ethical Committee of the District of Zürich (BASEC-Nr.: 2023-00110). Immediately following donation, all placental vessels were flushed with heparinized saline to prevent occlusions. A sheath was inserted and secured in the umbilical artery. Throughout the experiment, the placenta was continuously perfused with heparinized saline at a rate of 50 ml/min through the trocar. A 7 Fr guide catheter (Medtronic) was introduced via the trocar. A roadmap was obtained by manually injecting 5 ml of iodine-based contrast agent (Ultravist 300, Bayer) through the catheter. A magnetic capsule was introduced through the guide catheter and locomotion was achieved by setting the current in a single Navion coil to 35 A. Both the roadmap acquisition and capsule locomotion were recorded using a monoplanar fluoroscope (Ziehm Vision FD, Ziehm).

***In vitro thrombotic clot dissolution***: Human blood clots were retrieved from a human placenta, donated with informed written consent and the approval of the Ethical Committee of the District of Zürich (BASEC-Nr.: 2023-00110). Blood-clotting was allowed to proceed for a few hours before injecting them into a three-dimensional Middle Cerebral Artery (MCA) silicon model (Swiss Vascular GmbH). The blood clot was subjected to continuous water flow of 35 cm/s at 37°C for two hours to ensure stable vessel occlusion. Blood clot dissolution was induced by injecting a rtPA loaded capsule at the occluded vessel branch. Thermal capsule dissolution was allowed to proceed by increasing the water temperature to 41 °C for 2 min. During clod lysis, the capsule was magnetically held in the location of the blood clot.

***In vivo studies:***
*Visibility confirmation:* The animal study was approved by the local Committee for Animal Experimental Research (Cantonal Veterinary Office Zurich, Switzerland) under the license number ZH072/2023. Pigs were intact females with body weights (BW) of 50 kg..
On the day of the experiment, animals were sedated with an intramuscular injection of ketamine (Ketasol®-100 ad us.vet.; Dr. E. Graeub AG, Berne, Switzerland; 15mg/ kg BW), azaperone (Stresnil® ad us.vet.; Elanco Tiergesundheit AG, Basel, Switzerland; 2mg/ kg BW), and atropine (Atropinsulfat KA 1%; Kantonsapotheke, Zurich, Switzerland; 0.05mg/ kg BW). Anesthesia was then induced by administration of propofol (Propofol ® Lipuro 1%; B. Braun Medical AG, Sempach, Switzerland; 1-2mg /kg BW) in an intravenous line introduced in the V. auricularis until the swallowing reflex was repressed and intubation was possible. Anesthesia was maintained with a combination of constant intravenous infusion of propofol (Propofol ® Lipuro 2 %; B. Braun Medical AG, Sempach, Switzerland; 3 mg/kg BW/h) and inhalation of isoflurane (Attane™ Isoflurane ad.us.vet.; Piramal Enterpr. India; 1.0-1.5%) in 70 % oxygen:air mixture using mechanical ventilation.. A 7F distal access catheter was inserted through a 8F sheath and advanced to the common carotid artery (CCA) under fluoroscopic guidance. Capsules were then introduced through the distal access catheter and released into the blood flow within the CCA. Fluoroscopic imaging was performed using a mono-planar fluoroscope (Allura Xper FD20, Philips N.V.) at 30 frames per second.

*Electromagnetic navigation study:*
The animal study was performed as a feasibility test within the frame of an acute ovine trial performing magnetic guided laser endoscopy and was approved by the local Committee for Experimental Animal Research (Cantonal Veterinary Office Zurich, Switzerland) under the license number ZH117/2023 in conformity with the European Directive 2010/63/EU of the European Parliament and the Council on the Protection of Animals used for Scientific Purposes, and the Guide for the Care and Use of Laboratory Animals procedure. Anesthesia was induced by i.v. injection of ketamine hydrochloride (Ketasol®-100 ad us.vet.; Dr. E. Graeub AG, Berne, Switzerland; 3 mg/kg BW) in combination with Midazolam (Dormicum®, Roche Pharma (Schweiz) AG, Reinach, Switzerland; 0.2 mg/kg BW) and Propofol (Propofol ®- Lipuro 1%, B. Braun Medical AG; Sempach, Switzerland; 2-5 mg/kg BW). Sheep were orotracheally intubated. After intubation anesthesia was maintained with 1-1.5% Isoflurane in oxygen/air mixture in conjunction with a continuous infusion pump administering propofol (Propofol®- Lipuro 1%, B. Braun Medical AG;

Sempach, Switzerland 2 mg/kg/h) as well as a continuous infusion pump administering Ketamine at an infusion rate of 8mg/kg/h (Ketasol®-100 ad us.vet.; Dr. E. Graeub AG, Berne, Switzerland15mg). Throughout the procedure the animals received a continuous intravenous infusion of fentanyl (Fentayl® Sintetica SA, Mendrisio.Switzerland; 10-15ug/kg/h). The electromagnetic navigation system (eMNS), consisting of two Navions (NanoFlex Robotics AG and Magnebotix AG) positioned 34 cm apart, was placed perpendicular to the operating bed. A custom Styrofoam support was used to elevate and center the sheep, ensuring optimal positioning within the magnetic workspace. Capsules were introduced using a release gripper in combination with a 14G needle to place two capsules into the cisterna magna. The successful release of the capsules was verified before beginning the navigation process.

The successful release of the capsules was confirmed before proceeding with the navigation process. For navigation using rotating magnetic fields, a 35 mT rotating magnetic field with a frequency of 0.2 Hz was applied. For navigation using magnetic gradients, a gradient of 0.33 T/m was applied. These fields and gradients successfully guided the capsules through the subarachnoid space, directing them towards the fourth ventricle, and subsequently returning them to the cisterna magna.

## Supplementary Text 1
**Numerical simulations of in-flow navigation**

Numerical simulations

To numerically investigate the navigation of magnetic capsules along a bifurcation, the flow and the capsule motion were numerically simulated using a computational fluid dynamics approach based on the finite-volume method (FVM).(*47, 48*) Given the average flow velocity of each experimental run (i.e., $0.65 < \bar{u}_f < 0.85$ m s$^{-1}$) and the associated Reynolds number (i.e., $3200 < Re < 4250$), the flow regime ranges from transition to turbulent. Velocity and pressure were calculated by coupling the continuity equation with the Reynolds-averaged Navier-Stokes (RANS) equation for an incompressible Newtonian fluid, given by(*49*)

$$\nabla \vec{u}_f = 0 \tag{1}$$

$$\frac{\partial(\rho_f \bar{u}_i)}{\partial t} + \frac{\partial}{\partial x_j}\left(\rho_f \bar{u}_i \bar{u}_j + \rho_f \overline{u'_i u'_j}\right) = -\frac{\partial \bar{p}}{\partial x_i} + \frac{\partial}{\partial x_j}\left[\mu\left(\frac{\partial \bar{u}_i}{x_j} + \frac{\partial \bar{u}_j}{x_i}\right)\right] \quad i,j = 1,2,3 \tag{2}$$

where $\rho_f$ is the fluid density, $\bar{u}$ is the averaged flow velocity, $u'$ is the fluctuating flow velocity component, and $p$ is pressure. Besides the continuity and momentum equations above, the equations for generation, convection, and destruction of turbulence, as well as those describing the laminar-turbulent transition are modelled using the $\gamma - Re_\theta$ model from Langtry and Menter,(*50, 51*) which has been shown to adequately capture the laminar-turbulent transition, and has undergone thorough validation and sensitivity studies in the mentioned works.

The 3D domain of the bifurcation (main channel length of 96 mm, daughter branches length and diameter of 46 mm and 5 mm, respectively, fig. S5A) was meshed using both structured and unstructured elements (fig. S5C-D), with denser arrangement near the walls ensuring non-dimensional distances to the first cell of $y^+ \approx 1$, for accurately resolving of the viscous sublayer.(*51, 52*) Mesh independence tests were conducted with meshes obtained by dividing the main channel of the bifurcation in 75, 250 and 750 elements, with similar approaches followed for its daughter branches. For comparing the different meshes, we plotted the flow velocity profile (fig. S6) obtained at different lengthwise positions (x/D = -50, -30 and 20, Fig. S5E) and the pressure gradient obtained along the main channel (fig. S6B), The results of these comparisons indicate that the mesh with 250 lengthwise elements (i.e. 250, 20 and 20 along the x-, y- and z-axes, respectively, fig. S5C-D; ~1.4M elements) produces results that do not differ significantly from those obtained with a denser mesh (fig. S6A-B), and was therefore selected for all the simulations in this work.

The fluid properties were set based on those of water (density = 998.3 kg m$^{-3}$ and viscosity = 0.001 kg m$^{-1}$ s$^{-1}$) and the boundary conditions were chosen to mimic the experimental conditions. The average flow velocity considered in each experiment was imposed as boundary condition in each simulation at 80 diameters upstream the bifurcation inlet (fig. S5E) to allow the full development of the flow velocity profile (fig. S6C). No-slip condition was assumed at the walls and a zero-gauge pressure was set at the outlets of both daughter branches.
A steady-state, double-precision, pressure-based solver considering velocity-pressure coupling was used with second-order discretization. Convergence was assumed when residuals were less than 10$^{-5}$, with stricter criteria producing similar results.

Capsule navigation

The capsule was modelled as a sphere and its motion was predicted using a Lagrangian approach.(*53*) One-way coupling was used to model the interactions between the flow and the capsule, given its ability to predict the effect of flow on objects, with low computational cost.(*53*) The capsule velocity and position were obtained by integration of the force balance equation describing the contribution of the capsule momentum and that of the various forces acting on it, over discrete time steps. The total (net) force $\vec{F}_T$ acting on the capsule has contributions from drag (due to the flow), gravity, buoyancy (due to the capsule-fluid density mismatch) and the magnetic force (due to the applied external magnetic field gradient):(*54*)

$$m_p \frac{\mathrm{d}\vec{u}_p}{\mathrm{d}t} = \underbrace{m_p \frac{\vec{u}_f - \vec{u}_p}{\tau_p}}_{\vec{F}_{Drag}} + \underbrace{m_p \frac{\vec{g}(\rho_p - \rho_f)}{\rho_p}}_{\vec{F}_{Gravity} - \vec{F}_{Buoyancy}} + \underbrace{V_p \cdot \nabla\vec{B} \cdot \vec{M}}_{\vec{F}_{Magnetic}} \quad (3)$$

where $m_p$ is the capsule mass, $\vec{u}_p$ is the capsule velocity, $t$ is time, $\tau_p$ is the capsule relaxation time, $\vec{g}$ is the gravitational acceleration (9.81 m s$^{-2}$), $\rho_p$ is the capsule density, $V_p$ is the capsule volume, $\nabla\vec{B}$ is the magnetic field gradient and $\vec{M}$ is the capsule magnetization.

Based on the capsule characterization data, the capsule was modelled as having a 1.4 mm diameter and a density of 3187.74 kg m$^{-3}$, with its magnetization being dependent on the applied magnetic field (~30 mT), as described by the magnetization curves obtained experimentally (fig. S7).

The capsules were assumed to be transported by the flow and enter the bifurcation at 20 different inlet radial positions (to account for different possible entrance locations into the main channel; (fig. S5B), with the flow velocity and direction prevailing at the mentioned positions. As an upgrade to the classical one-way approach where objects are treated as points (i.e. the capsule center) rather than volumes, in this work, we acknowledged collisions with the walls when the center of the capsules was one radius away from the walls. This prevented the capsules from moving beyond the walls (as it can happen when the capsules are treated as volume-less objects, as in the classical one-way approach) and generated more accurate predictions of the capsule's trajectories.(55) Furthermore, the capsules were assumed to undergo fully elastic collisions with the walls, and thus to retain their momentum upon collision.

For computing the capsule trajectories, a time step of $\Delta t = 10^{-5} \cdot (u_p^i + u_f^i)^{-1}$ was used for the iteration of equation (3), with shorter steps yielding similar results (fig. S6D). At each iteration of the capsule position, the magnetic field and the magnetic field gradient measured at the workspace where the bifurcation was positioned during the experiments, was interpolated to obtain the field and the gradient prevailing at every capsule position along the bifurcation. At every position, the field was used to calculate the capsule magnetization (fig. S7) and the magnetic gradient was used to compute the magnetic force acting on the capsule (Equation (3)). This iterative process was repeated until the capsule exited through one of the bifurcation outlets.

Validation of the modelling approach

To confirm the validity of the predictions obtained with the $\gamma - \mathrm{Re}_\theta$ transition model used in this work, we conducted two different comparisons. First, we numerically replicated the T3A flat plate experimental case from the T3 series of experimental tests reported in the literature for use in the validation of transition models.(56) We compared the skin friction coefficient predicted along the surface of the flat plate when using the $\gamma - \mathrm{Re}_\theta$ transition model, to that obtained in the experimental test (fig. S8), and observed a good match between the predictions and the experimental data. This agreement shows that the modelling approach followed in this work can be used to predict different flows across the laminar-turbulent regime.

Finally, we calculated the friction factor for the bifurcation for Re ≈ 4250, and compared it to the value retrieved from the classical Moody chart for smooth pipes ($f \approx 0.04$).(52) In the bifurcation main channel (x = 0 to 96 mm, fig. S5E) the flow is fully developed, and thus, the friction factor is given by(52)

$$f = \Delta p \frac{D}{\frac{1}{2}\rho\bar{u}^2 L} = 279.03 \cdot \frac{0.005}{\frac{1}{2} \cdot 998.3 \cdot 0.85^2 \cdot 0.096} = 0.0403 \quad (4)$$

where $\Delta p$ is the pressure drop predicted to occur along the main channel of the bifurcation given the associated length ($L$), diameter ($D$) and average flow velocity ($\bar{u}$). That the predicted friction factor ($f \approx 0.0403$, Equation (4)) is very similar to the value retrieved from the Moody chart for smooth pipes ($f \approx 0.04$), is a further indicator that turbulent flows are accurately predicted by the present modelling approach.

Cases considered in the experiments and simulations

A full factorial design-of-experiments approach was used to prepare the set of experimental conditions to experimentally test and numerically replicate using the present modelling approach. Five different average flow velocities (0.65 - 0.85 m·s$^{-1}$) and ten different magnetic gradient magnitudes (0 - 450 mT·m$^{-1}$, Table S1) were combined in a full factorial approach, where each possible combination of the two parameters generated an independent test case. For each pair of average flow velocity and magnetic field gradient (Table S1), 18-19 experiments were conducted where a capsule was flowed through the bifurcation (fig. S5A) under the effect of a constant magnetic field (~30 mT) and a constant magnetic gradient, applied perpendicularly to the bifurcation main channel. In the simulations replicating the experiments, the capsule was released in the bifurcation at 20 different radial positions (fig. S5B), for a total of 1000 simulations (i.e. 20 capsule positions × 5 flow velocities × 10 magnetic gradient values). For each experiment / simulation, the success of the magnetic navigation was assessed by calculating the ratio between the number of capsules reaching the desired bifurcation outlet and the total number of capsules entering the bifurcation in the first place.

Results

Fig. S9 shows the success of the magnetic navigation obtained for each pair of flow velocity and magnetic gradient, observed in the experiments and predicted in the simulations. In line with the underlying physics of the problem, the success of navigation generally increases with increasing magnetic gradient (due to the increasing magnetic forces induced in the capsule) and decreasing flow velocity (due to the larger time window to actuate the capsule, fig. S9). These effects are seen in both the experimentally determined navigation success, and that predicted by numerical simulation (left and right tables in fig. S9, respectively). More importantly, the predicted navigation success follows the obtained experimental values, both in the direction of variation, and in the absolute values of navigation success, even if some small discrepancies do exist. In this regard, when comparing the predicted navigation success with the experimental data, we see that the predicted number of capsules reaching the desired outlet differed by no more than 1-4 capsules, out of the 20 considered for each simulation case. Note that these small differences may result from a small mismatch between the conditions prevailing in the experiments and those assumed in the simulations, e.g. regarding the momentum conservation upon capsule collision with the walls, or the capsules exact position, velocity and direction when entering the bifurcation. Note that any such mismatch may have resulted in slightly different trajectories and collisions with the walls, that may have led to diversion of some of the capsules to the undesired outlet. Note also that the downward pull associated with the gravitational acceleration very likely induced, both in the experiments and in the simulations, collisions with the bifurcation lower walls that may have affected the trajectories of the capsules in complex ways due to the non-flat shape of the bifurcation walls. This and the other effects mentioned above influence the trajectory of the capsules along the bifurcation and, ultimately, the navigation success that is measured in the experiments and that predicted by the simulations.

Nevertheless, the fact that the simulations of the capsule motion along the bifurcation under magnetic manipulation generated navigation success predictions following the values obtained experimentally, for a wide set of flow velocity and magnetic gradients, indicates that the forces accounted for in the simulations (i.e. drag, gravity, buoyancy and magnetic force) are indeed the key forces influencing the trajectory of the capsules. More importantly, the results obtained in the experiments and in the simulations show that the capsules can be magnetically navigated along bifurcations, and that, for any of the flow velocities considered in this work (relevant for various clinical applications), it is possible to actuate the capsule with a strong enough magnetic force for ensuring successful navigation. This offers a multitude of possibilities in terms of capsule manipulation, that can be explored for developing innovative clinical interventions.

# Supplementary Text 2

**Pharmacological Evaluation**

We characterized the drug-loading capacity of the microrobot and evaluated its ability for long-term release. The model drugs Doxorubicin (DOX) and Ciprofloxacin (CIPRO) were selected due to their widespread application in chemotherapeutic and antibiotic treatments, respectively. These small-molecule drugs were incorporated into the capsules via diffusion, achieved by incubating the capsules in highly concentrated aqueous solutions of each therapeutic agent. The drug release profiles of DOX- and CIPRO-loaded capsules revealed a pH-dependent release mechanism (fig. S10A–B), enabling controlled and timed therapeutic delivery tailored to the pH of the surrounding surrounding tissue environment.

To assess the system's capacity for delivering large molecules, we evaluated its compatibility with recombinant tissue plasminogen activator (rtPA), the gold standard for systemic acute ischemic stroke therapy. We observed a natural tendency of rtPA to adsorb onto the surface of the nanoparticles (fig. S10C-D) embedded in the microrobot, resulting in a reduction in overall enzymatic activity to 58%. To examine the long-term stability of rtPA, particularly its autocatalytic conversion from single-chain rtPA (sc-rtPA) to the higher active form two-chain rtPA (tc-rtPA), we monitored the activity of rtPA and the sc-rtPA contend over a three-month period (fig. S10E-F). According to the European Pharmacopoeia quality standards, at least 60% of rtPA-related substances must consist of sc-rtPA, a condition achievable for three months by storing the compound at 5 °C. Further, to evaluate rtPA stability in a gelatin matrix, we conducted a long-term study of rtPA activity in a 5% (w/v) gelatin solution (fig. S10G). Similar to storage in water, samples stored at 5°C did not exhibit increased enzymatic activity indicating no conversion to tc-rtPA. However, samples stored at 25°C showed a 1.7-fold increase in activity, though this was less pronounced compared to the 3.1-fold increase observed in water, suggesting a stabilizing effect of the gelatin matrix.

The stability of rtPA at different temperatures was further studies by incubating rtPA solutions at 37 °C, 50 °C, 60 °C, and 70 °C for 90 minutes (fig. S10H). To ensure that the hyperthermia used for capsule dissolution did not compromise the rtPA stability, we dissolved the microrobots in water at 38°C and used hyperthermic dissolution. No significant difference in rtPA activity was detected between the two methods, indicating that hyperthermic dissolution did not affect the rtPA stability (fig. S10I).

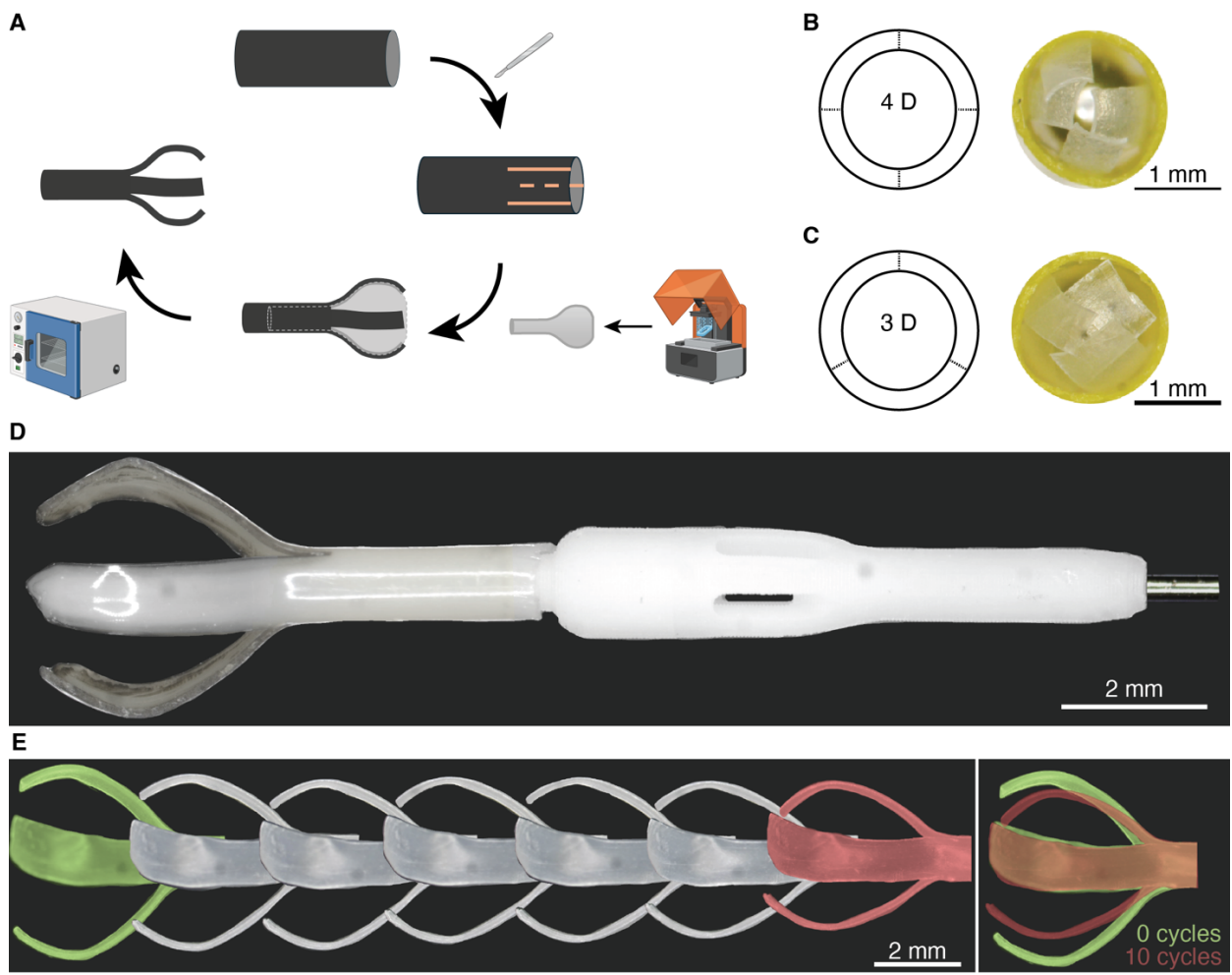

**Fig. S1. Release gripper fabrication and characterization.** (A) Fabrication scheme of the release gripper. (B) Schematic representation of a four-digit release gripper design. (C) Schematic representation of a three-digit release gripper design. (D) Optical microscope image of an assembled release gripper. (E) Deformation analysis of the release gripper after 10 loading cycles, demonstrating structural reliability.

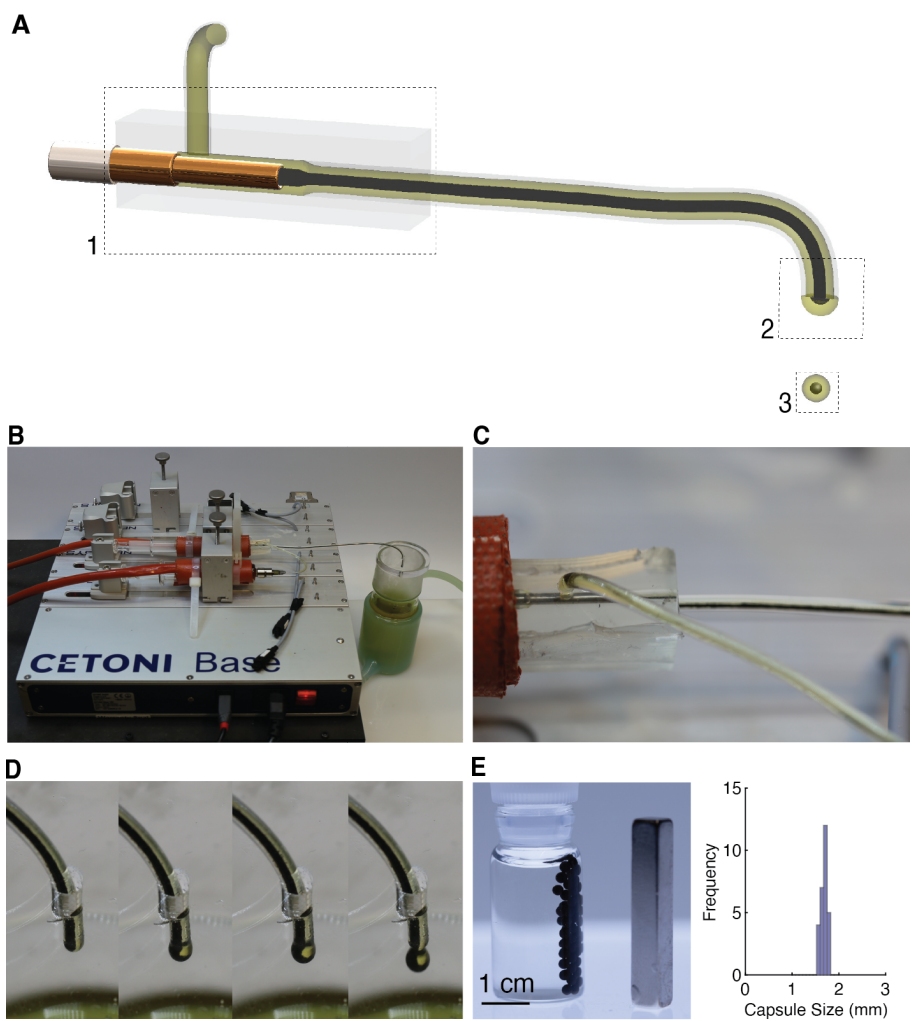

**Fig. S2. Microfluidic capsule fabrication process.** (A) Schematic representation of the microfluidic 3D flow-focusing chip: (1) Generation of a stable co-flow of the capsule hydrogel and oil, (2) droplet formation at the outlet, and (3) droplet falling cooled oil. (B) Image of the capsule fabrication setup. (C) Close-up image of the stable co-flow within the chip. (D) Image series illustrating the droplet formation process. (E) Image of multiple fabricated capsules (left) and their size distribution (right).

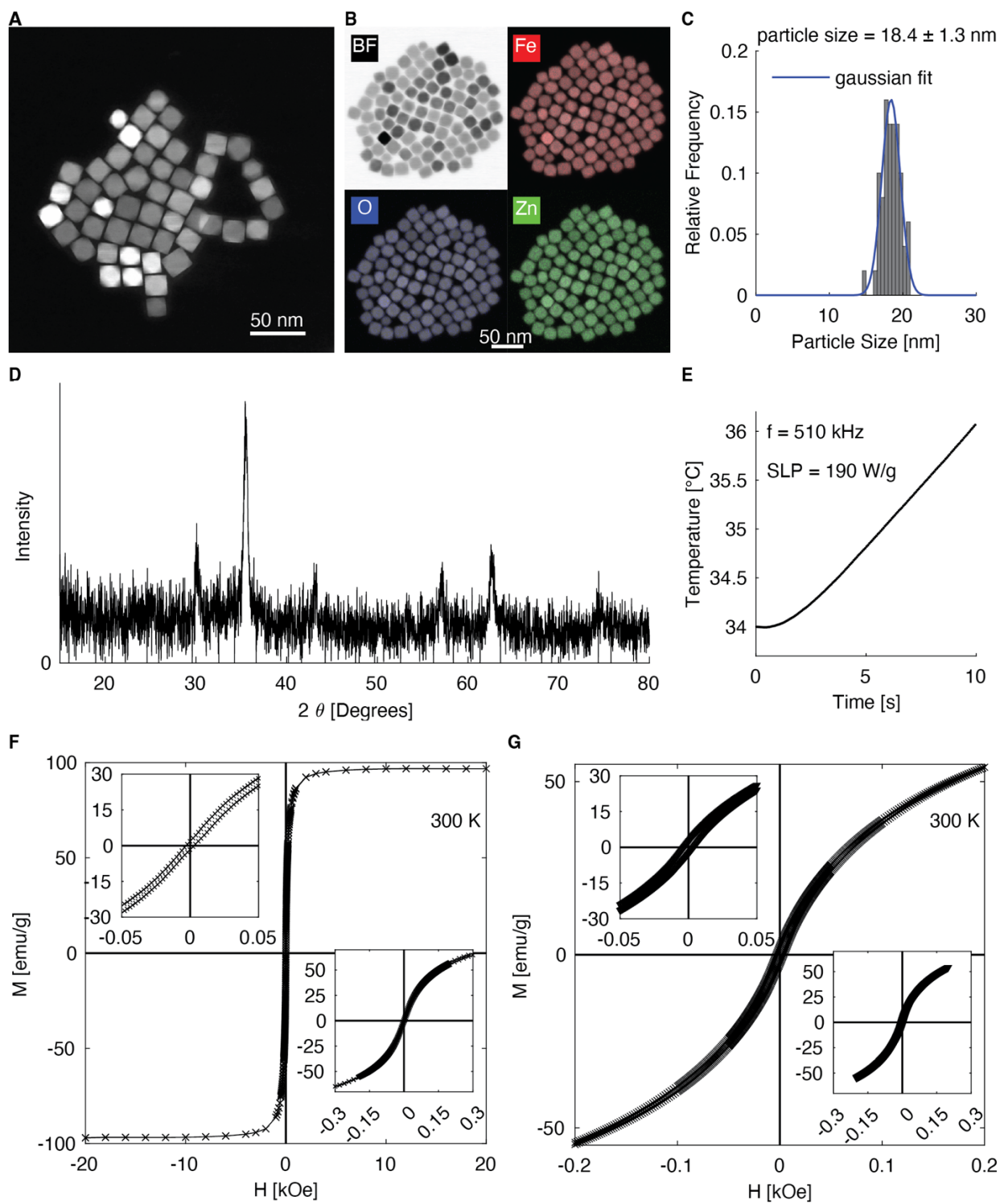

**Fig. S3. Characterization of zinc-substituted iron oxide nanocubes.** (A) TEM image displaying uniform zinc-substituted iron oxide nanocubes. (B) Elemental mapping using EDX confirming the presence of Fe, Zn, and O in the nanocubes. (C) Particle size distribution fitted with a Gaussian curve. (D) XRD pattern validating the inverse spinel structure of the nanocubes. (E) Hyperthermic heating profile under an alternating magnetic field (510 kHz at 20 mT), with a specific loss power

(SLP) of 190 W/g. (F) 2 T magnetic hysteresis loop and 300 K. (G) 20 mT magnetic hysteresis loop at 300 K.

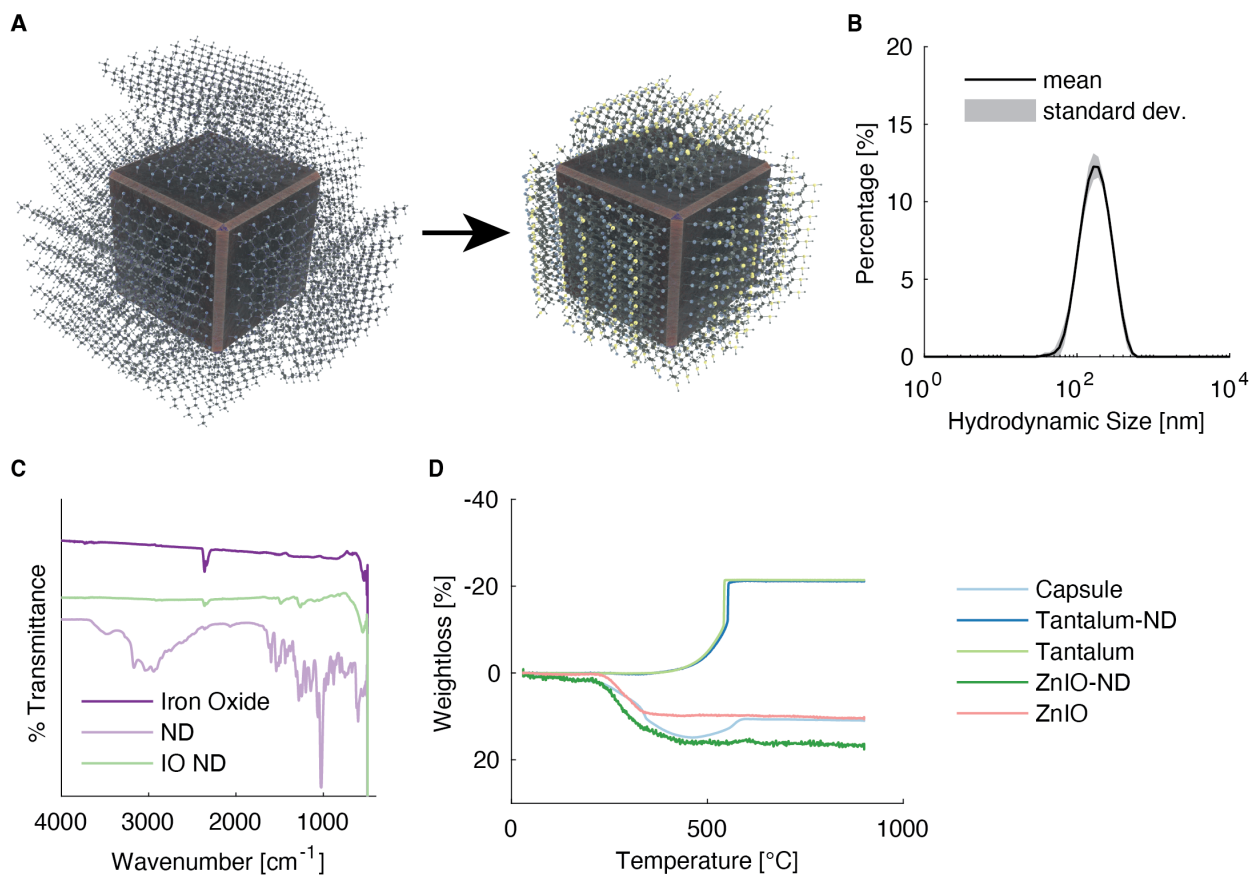

**Fig. S4. Ligand exchange characterization.** (A) Schematic representation of the ligand exchange process on iron oxide nanocubes. (B) Particle size distribution showing mean and standard deviation after ligand exchange. (C) FTIR spectra of iron oxide nanocubes, nitrodopamine, and nitrodopamine-coated iron oxide nanocubes. (D) Thermogravimetric analysis (TGA) comparing weight loss profiles of the capsule, nitrodopamine-coated tantalum particles, pure tantalum particles, nitrodopamine-coated iron oxide nanocubes, and pure iron oxide nanocubes.

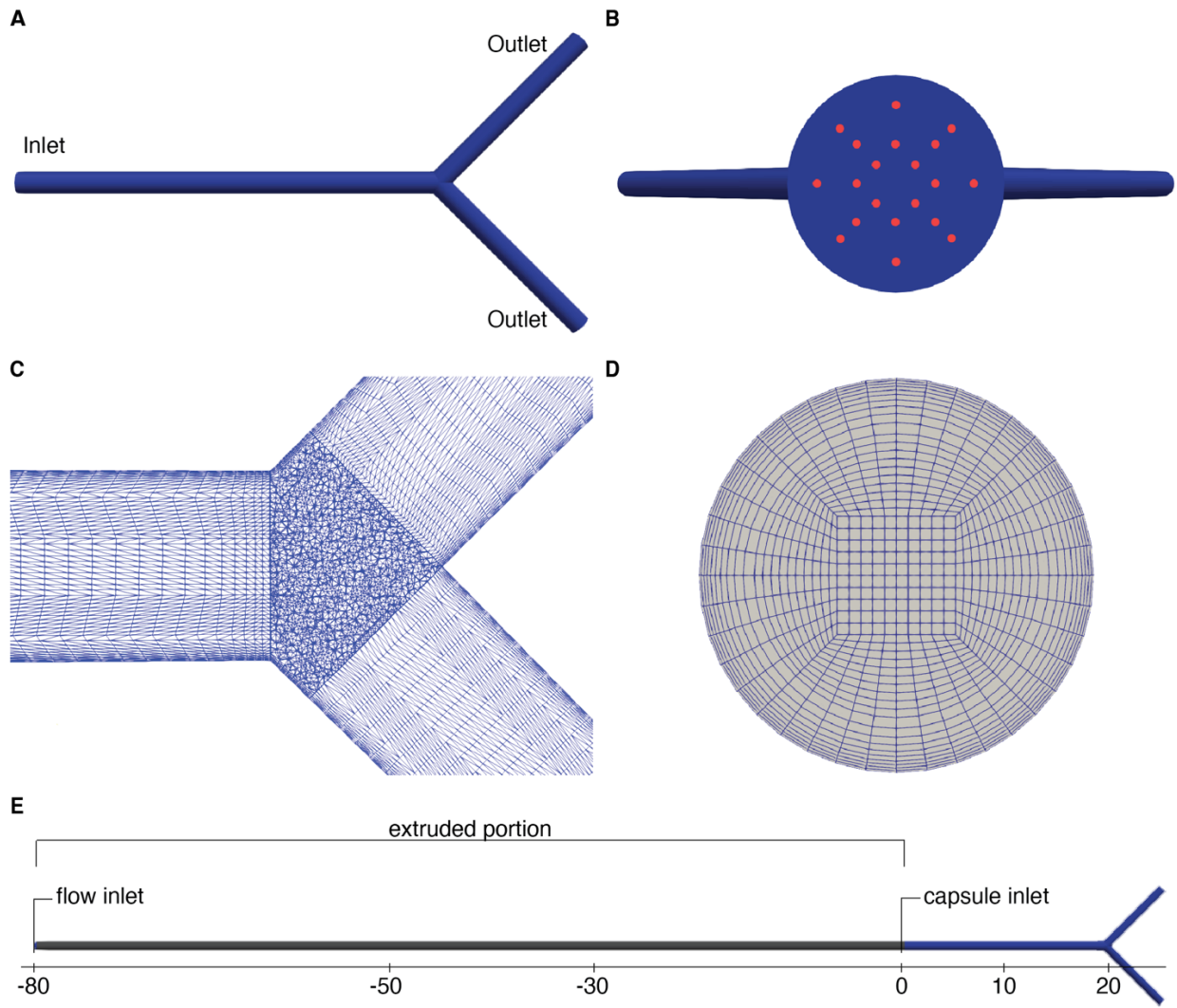

**Fig. S5. Geometry, mesh and entrance positions of the capsules, considered for the numerical simulations.** A) 3D geometry of the bifurcation with main channel length of 96 mm (extruded section not shown), daughter branches length and diameter of 46 mm and 5 mm, respectively; B) 20 entrance positions of the capsules entering the bifurcation; C) and D) Mesh details of the bifurcation and inlet cross-sections, respectively, highlighting the use of structured and unstructured elements; and E) Representation of the 3D geometry considered in the simulations, including the bifurcation, the extruded portion upstream its inlet, and the relevant positions in the x-axis considered in the mesh independence studies of fig. S6.

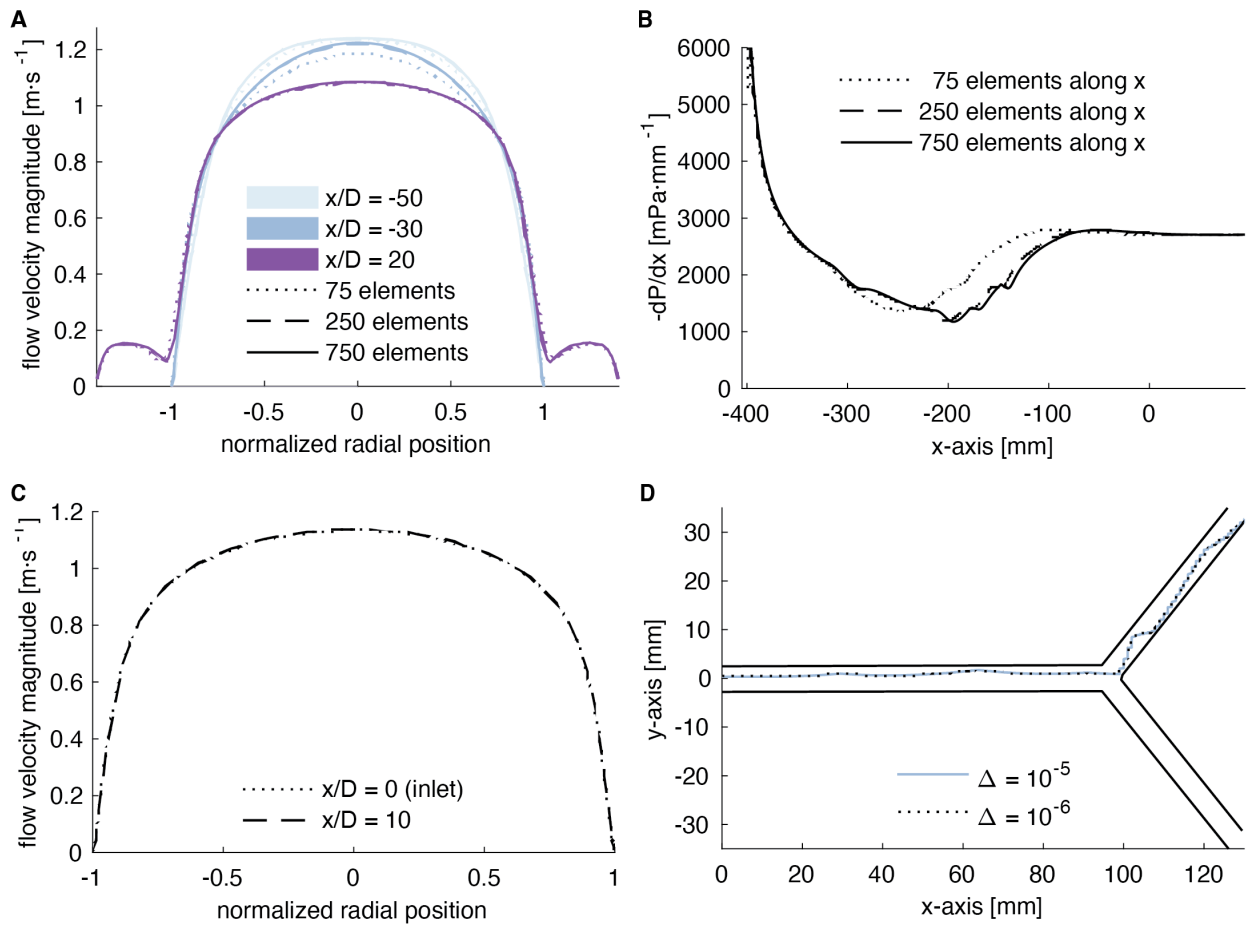

**Fig. S6. Mesh independence tests for the flow velocity magnitude, the pressure gradient and the capsule time-step.** A) Velocity profiles at three different positions (x/D = -30, -50 and 20; fig. S5E) for three different meshes (with 75, 250 and 750 elements along the bifurcation main channel); B) Pressure gradient along the bifurcation for the three meshes considered. The results from A) and B) show that the mesh using 250 elements is sufficient, with denser meshes yielding similar results; C) Flow velocity profile at two different positions (x/D = 0 and 10; fig. S5E) that show the fully developed profile along the bifurcation; and D) Capsule trajectories using a time step of $10^{-5}$ and $10^{-6}$, showing that a time step of $10^{-5}$ is sufficient and that a lower time step produces a similar trajectory.

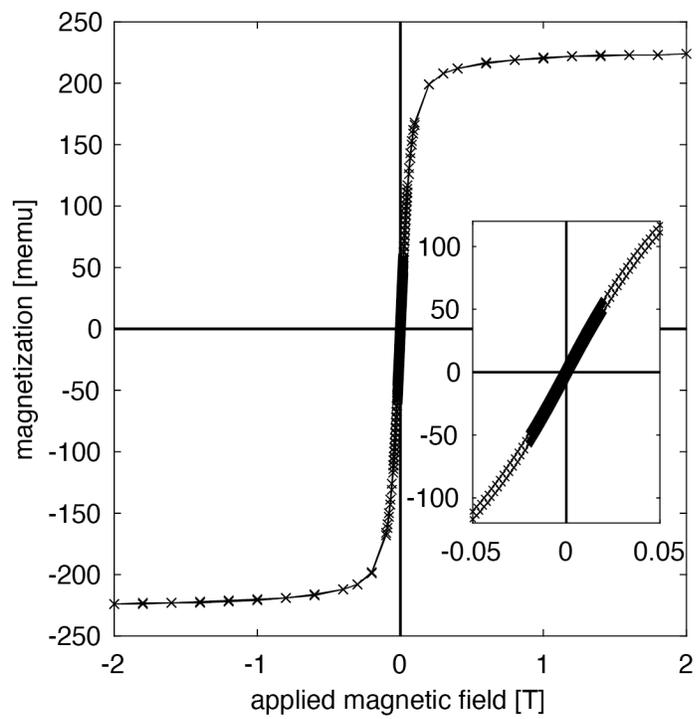

**Fig. S7.** Magnetization curve obtained for the capsules used in the experiments.

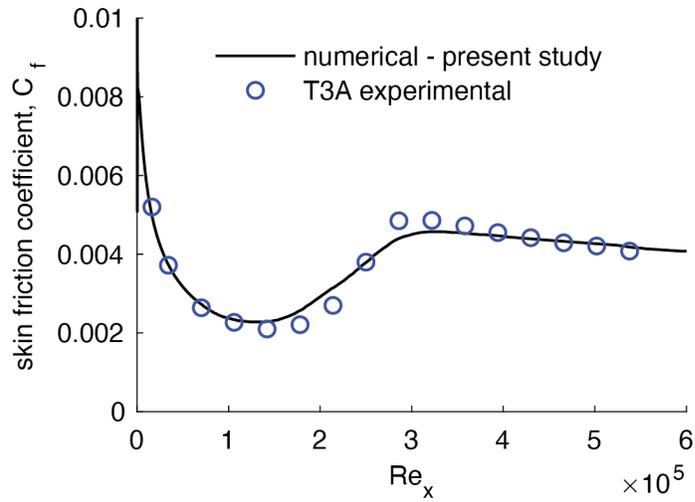

**Fig. S8. Validation of Transition Model: Skin Friction Coefficient.** Validation of the numerical transition model used in our numerical simulations by comparison of the skin friction coefficient obtained numerically and experimentally for the T3A flat plate test, The friction coefficient is plotted along the plate surface, represented here by the local Reynolds number ($Re_x$), to highlight the variation in the coefficient due to the laminar-turbulent transition.

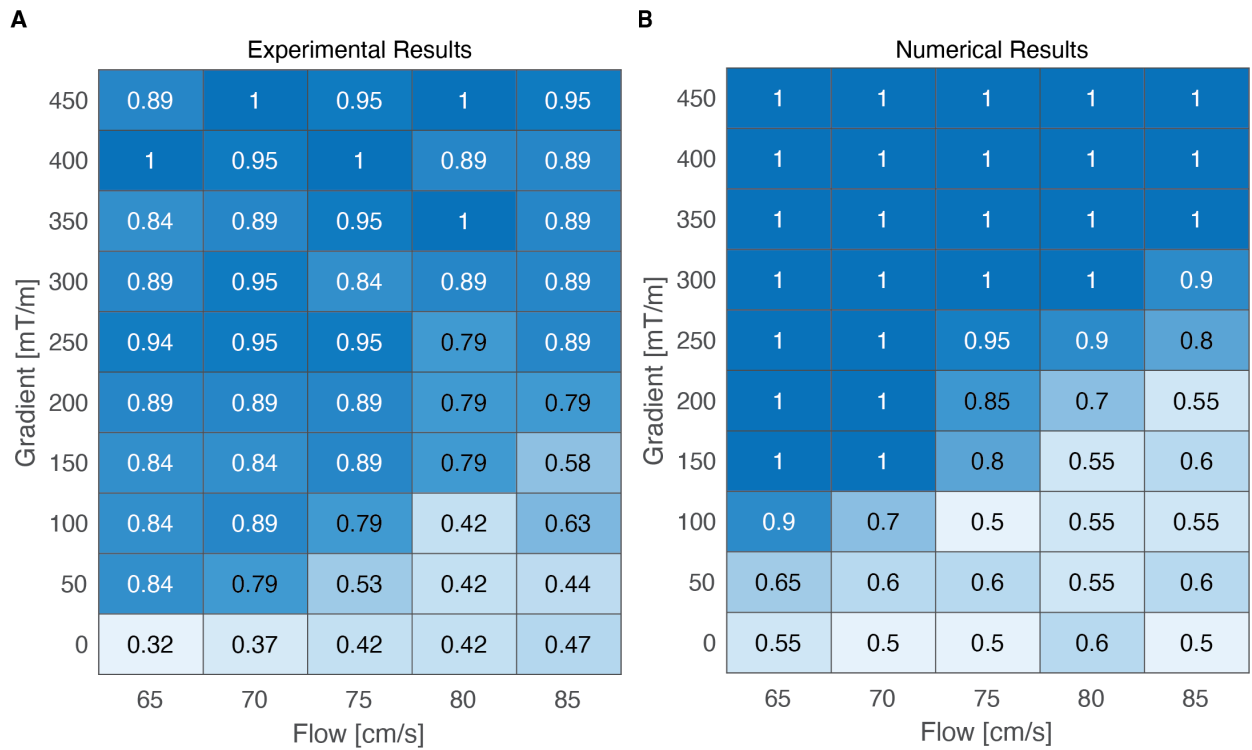

**Fig. S9. Magnetic Navigation Success: Experimental vs. Predicted Results.** Experimental and predicted magnetic navigation success calculated as the ratio between the number of capsules reaching the desired bifurcation outlet and the total number of capsules introduced in the bifurcation (18-19 in the experiments and 20 in the simulations, one at a time). These results were obtained considering different magnetic gradient and average velocity values for both the experiments and numerical simulations and show that the modelling approach can adequately predict the magnetic navigation. The slight discrepancies observed in the number of capsules reaching the outlet (i.e., 1-4 capsules) may be due to a mismatch between the capsules entrance position, velocity and direction as well as the elasticity of the capsule-wall interactions, assumed in the numerical simulations and those prevailing in the experiments.

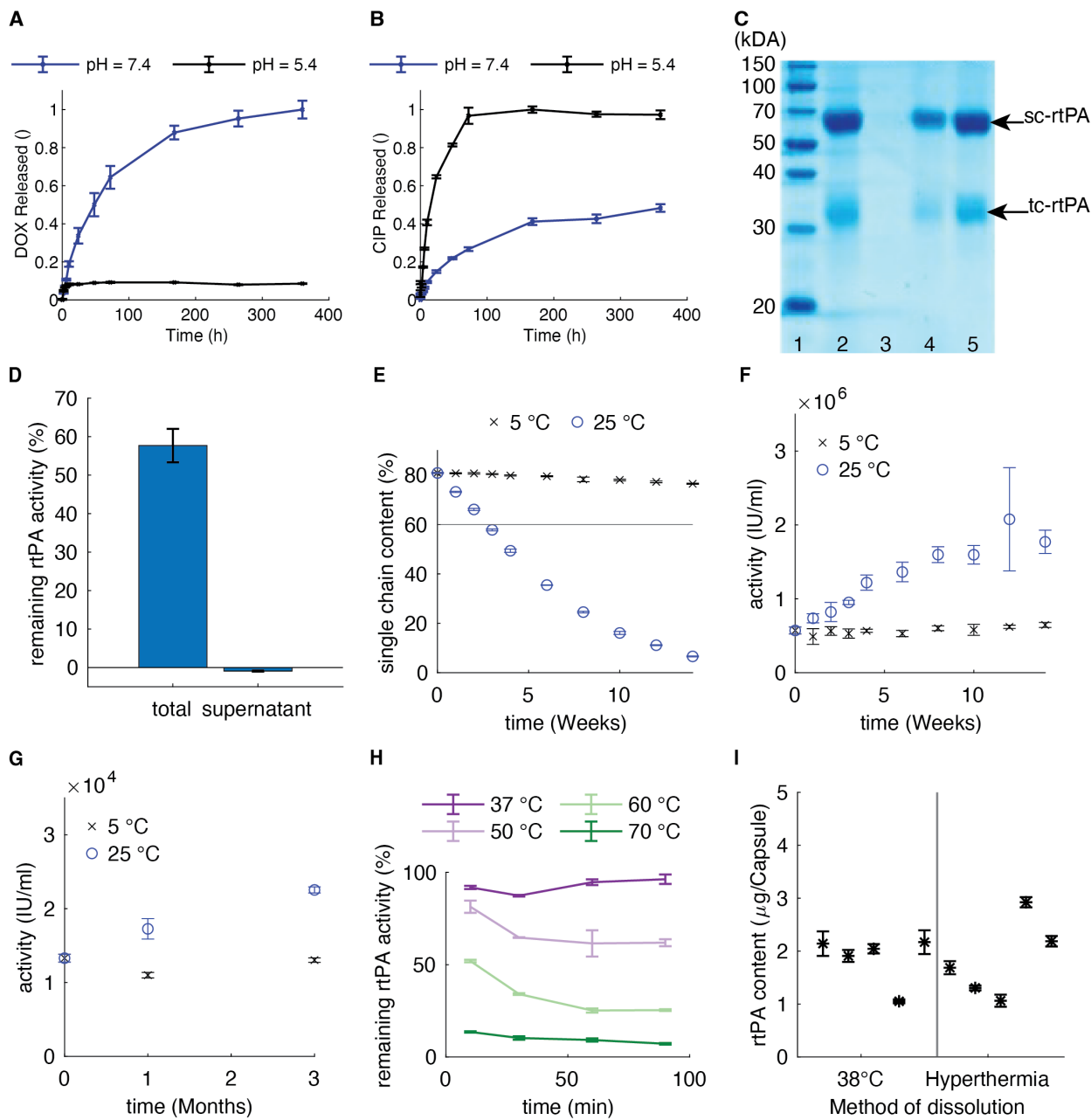

**Fig. S10. Investigation of therapeutic content and stability.** (A) pH-dependent long-term drug release of Doxorubicin (DOX) from the capsules. (B) pH-dependent long-term drug release of Ciprofloxacin (CIPRO) from the capsules (C) 12% SDS-Page under reductive conditions. 1 Protein marker in kDa, 2 rtPA with centrifuged supernatant of ION suspension, 3 rtPA with ION suspension incubated for one hour, 4 rtPA with ION suspension without incubation, 5 rtPA reference solution (sc-rtPA: 63.6 kDa, tc-rtPA: two bands 30-35 kDa). (D) Remaining enzyme activity of rtPA after incubation with IONs for one hour. (E) Single-chain content determined by size exclusion chromatography of rtPA stored in water for injection at 5°C (cross) and 25°C (circle) for 14 weeks. Dashed horizontal line marks the quality requirement of the European Pharmacopoeia of 60% sc-rtPA content. (F) Enzyme activity of rtPA determined by fluorescence activity assay of rtPA stored in water for injection at 5°C (cross) and 25°C (circle) for 14 weeks. (G) Enzyme activity of rtPA determined by fluorescence activity assay of rtPA stored in 5% (m/V)

gelatin solution at 5°C (cross) and 25°C (circle) for 3 months. (H) Enzyme activity of rtPA after incubation at 37 to 70°C for 90 minutes. (I) Enzyme activity of rtPA per microrobotic capsule determined by fluorescence activity assay after using 38°C or hyperthermia as dissolution methods. Each dissolution method was performed with 5 capsules.

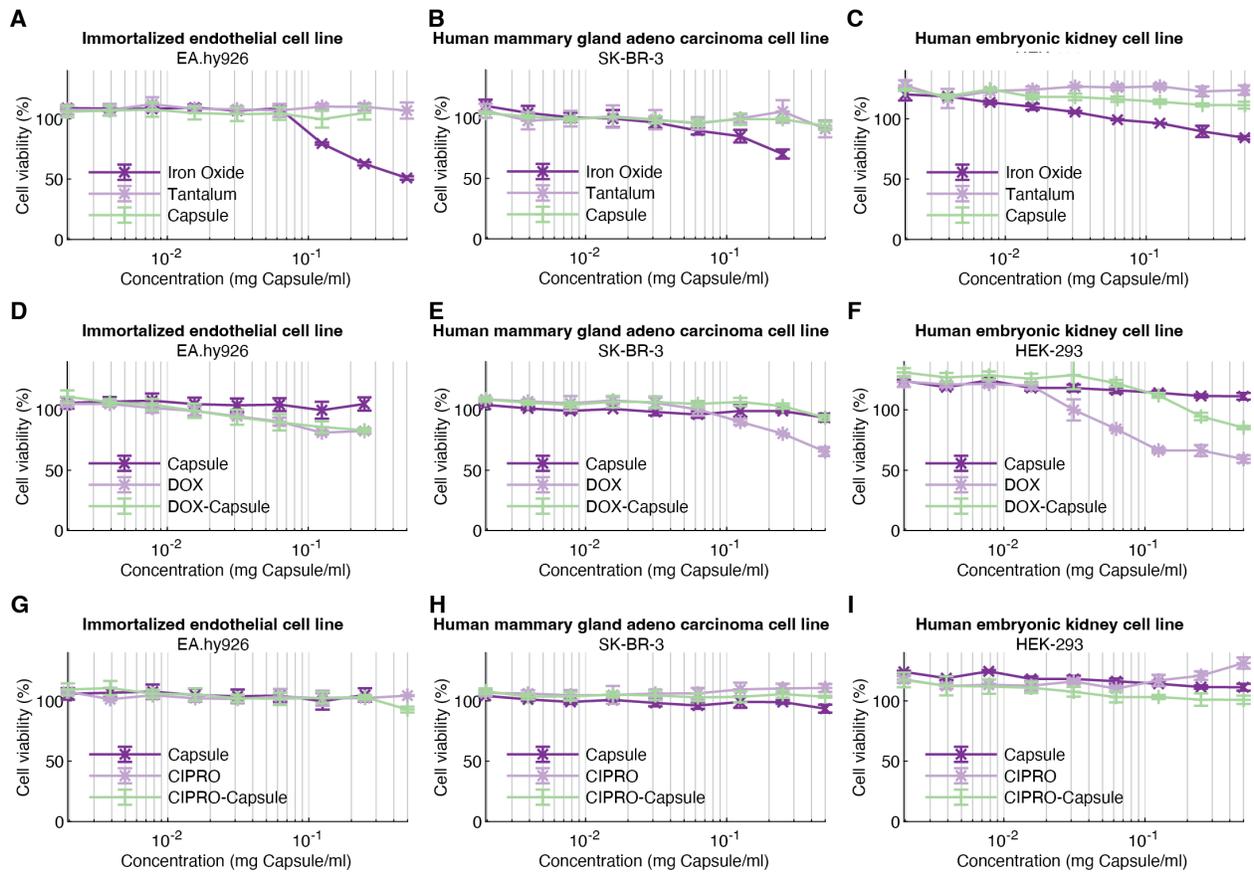

**Fig. S11. Cell viability studies.** Influence of (A-C) microrobotic capsule ingredients, (D-F) doxorubicin microcapsules and (G-I) ciprofloxacin microcapsules on cell viability of EA.hy 926, SK-BR-3 and HEK 293 cell lines. The investigated concentrations of the pure ingredients: Ion Oxide, Tantalum, Cirpofloxacin and Doxorubicin, correspond to the concentrations in the respective capsule formulation in mg capsule/mL. Empty capsules, Ciprofloxacin capsules and Doxorubicin capsules are the capsule formulations. Mitochondrial activity was assessed after 24 hours using WST-1 reagent.

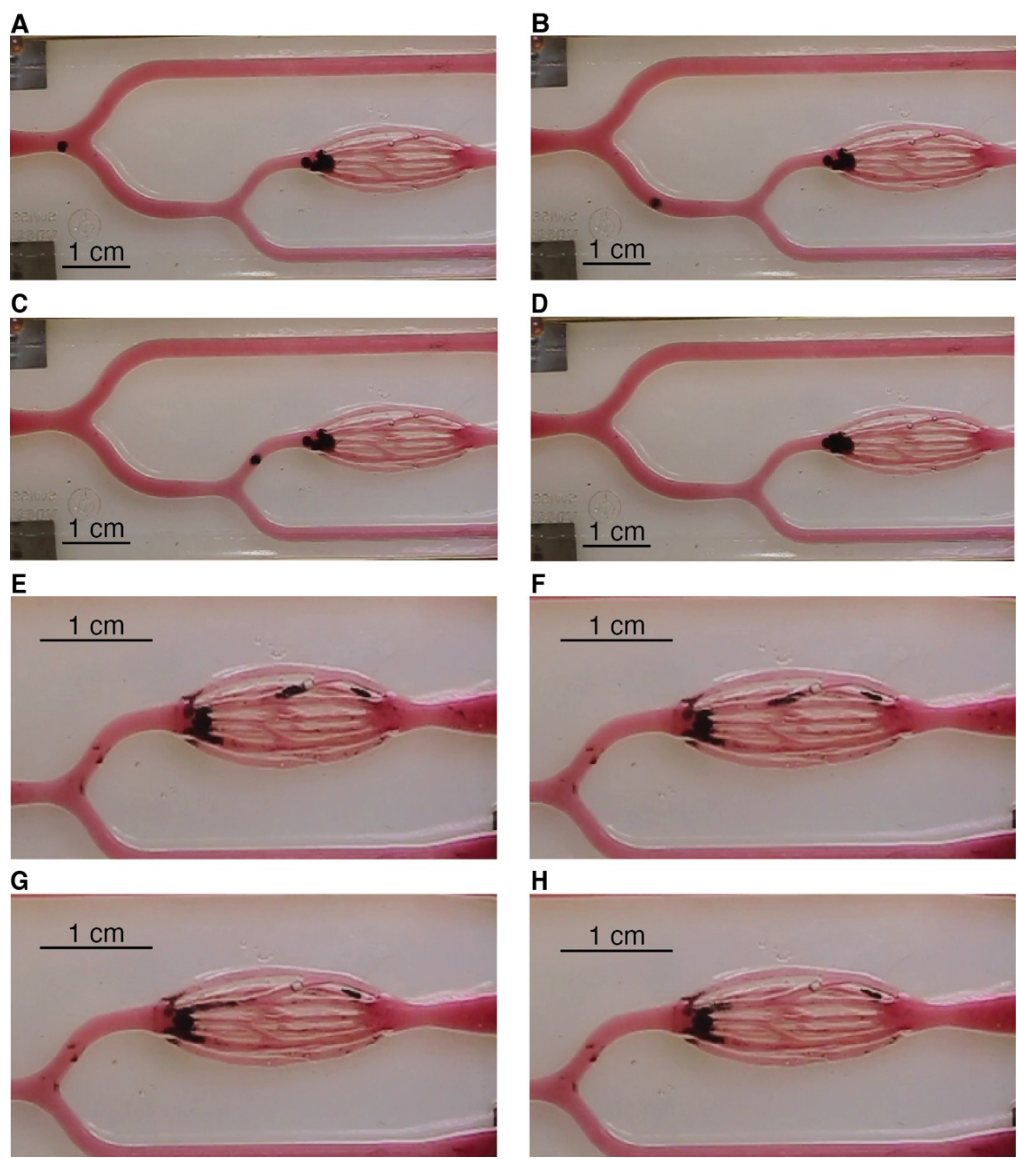

**Fig. S12. Magnetic navigation and nanoparticle swarm formation for microvasculature navigation in a silicone model.** (A-C) Magnetic navigation of microrobots toward a microvascular structure. (D) Embolization of the microvascular structure upon microrobot arrival. (E) Dissolution of the capsules and release of nanoparticle swarms. (F-G) Locomotion of the nanoparticle swarms against the flow, moving from right to left.

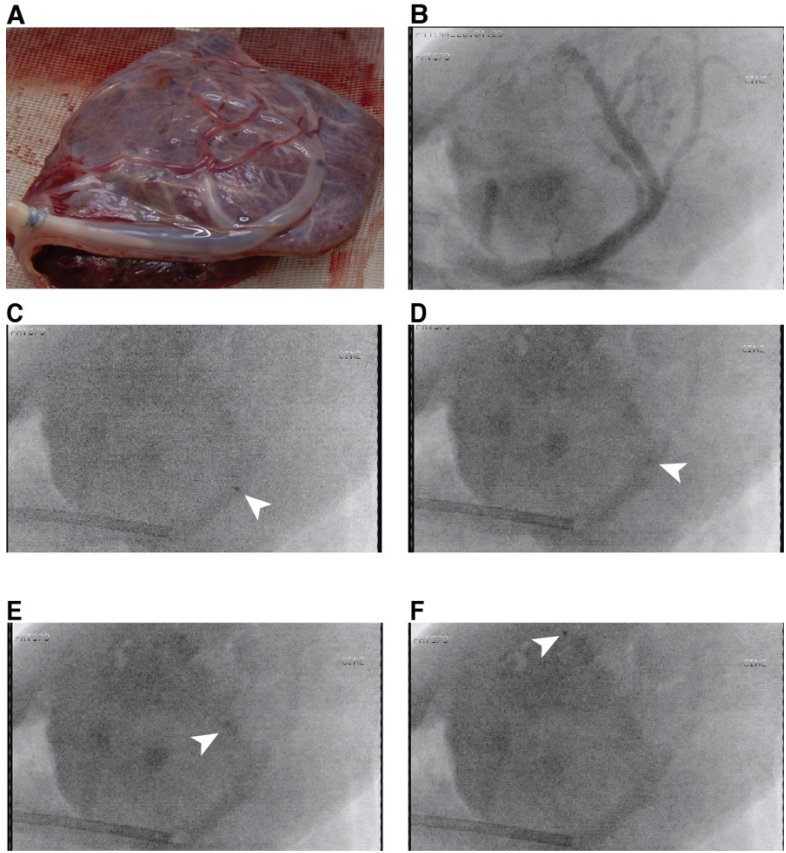

**Fig. S13. Capsule navigation *ex vivo in* a human placenta. (**A) Image of the human placenta. (B-F) Fluoroscopic views showing the capsule being magnetically guided through the main vessel of the placenta, demonstrating precise navigation and visibility under fluoroscopic imaging.

**Table S1.**

Average flow velocities and magnetic gradient magnitudes considered for the experiments and the numerical simulations. These conditions were combined following a full factorial design-of-experiments, with 18-19 experiments and 20 simulations considering the release of the capsules at different radial positions (Figure 1B), being performed per each pair of flow velocity and magnetic gradient. 1000 simulations were run (20 capsule positions × 5 flow velocities × 10 magnetic gradients).

| average flow velocity [m·s$^{-1}$] | magnetic gradient [mT·m$^{-1}$] |
| --- | --- |
| 0.65 | 0 |
| 0.70 | 100 |
| 0.75 | 150 |
| 0.80 | 200 |
| 0.85 | 250 |
|  | 300 |
|  | 350 |
|  | 400 |
|  | 450 |